\documentclass{article}
\usepackage{arxiv}

\usepackage[utf8]{inputenc} 
\usepackage[T1]{fontenc}    
\usepackage{hyperref}       
\usepackage{url}            
\usepackage{booktabs}       
\usepackage{amsfonts}       
\usepackage{nicefrac}       
\usepackage{microtype}      
\usepackage{lipsum}
\usepackage{float}
\usepackage{graphicx}
\usepackage{multirow}
\usepackage{array}
\usepackage{comment}
\usepackage{longtable}
\usepackage{ragged2e}
\usepackage{subfig}
\usepackage[table]{xcolor}  
\usepackage{geometry}
\usepackage{multicol}
\usepackage{amsmath}
\usepackage{pifont}
\title{A Comprehensive Survey on Diffusion Models and Their Applications}

\author{
  Md Manjurul Ahsan \\
  Industrial and Systems Engineering\\
  University of Oklahoma\\
  Norman, Oklahoma-73071 \\
  \texttt{ahsan@ou.edu} \\
   \And
  Shivakumar Raman \\
  Department of Industrial and Systems Engineering\\
  University of Oklahoma\\
  Norman, Oklahoma-73071\\
  \texttt{raman@ou.edu} 
   \And 
  Yingtao Liu \\
  Department of Aerospace and Mechanical Engineering\\
  University of Oklahoma\\
  Norman, Oklahoma-73071\\
  \texttt{yingtao@ou.edu} 
   \And
  Zahed Siddique \\
  Department of Aerospace and Mechanical Engineering\\
  University of Oklahoma\\
  Norman, Oklahoma-73071\\
  \texttt{zsiddique@ou.edu} 
}

\begin{document}
\maketitle

\begin{abstract}
\textbf{Diffusion Models} are probabilistic models that create realistic samples by simulating the diffusion process, gradually adding and removing noise from data. These models have gained popularity in domains such as image processing, speech synthesis, and natural language processing due to their ability to produce high-quality samples. As \textbf{Diffusion Models} are being adopted in various domains, existing literature reviews that often focus on specific areas like computer vision or medical imaging may not serve a broader audience across multiple fields. Therefore, this review presents a comprehensive overview of \textbf{Diffusion Models}, covering their theoretical foundations and algorithmic innovations. We highlight their applications in diverse areas such as media quality, authenticity, synthesis, image transformation, healthcare, and more. By consolidating current knowledge and identifying emerging trends, this review aims to facilitate a deeper understanding and broader adoption of \textbf{Diffusion Models} and provide guidelines for future researchers and practitioners across diverse disciplines.

\end{abstract}


\keywords{Diffusion Models\and Generative Modeling\and Synthetic Data Generation\and Image Synthesis\and Image-to-Image Translation\and Text-to-Image Generation\and Audio Synthesis\and Time Series Forecasting\and Anomaly Detection\and Medical Imaging\and Data Augmentation\and Computational Efficiency\and Uncertainty Quantification\and Riemannian Manifolds\and Molecular Dynamics\and Super-Resolution\and Semantic Image Synthesis\and Zero-Shot Classification\and Atmospheric Turbulence Correction}

\section*{Abbreviations}

\begin{multicols}{2}
\begin{itemize}
    \item \textbf{ACDM:} Autoregressive Cascade Multiscale Diffusion
    \item \textbf{ACDMSR:} Accelerated Conditional Diffusion Model for Image Super-Resolution
    \item \textbf{AIoT:} Artificial Intelligence of Things
    \item \textbf{BerDiff:} Bernoulli Diffusion Model
    \item \textbf{BLIP:} Bootstrapped Language-Image Pretraining
    \item \textbf{BuilDiff:} Building Diffusion
    \item \textbf{CDDM:} Conditional Denoising Diffusion Model
    \item \textbf{CDMs:} Classifier-guided Diffusion Models
    \item \textbf{CIDEr:} Consensus-based Image Description Evaluation
    \item \textbf{CLE:} Controllable Light Enhancement Diffusion
    \item \textbf{CLIP:} Contrastive Language-Image Pre-training
    \item \textbf{CLIPSonic:} Controlled Language-Image Pretraining Sonic
    \item \textbf{CMD:} Conditional Diffusion Models
    \item \textbf{DDIM:} Denoising Diffusion Implicit Models
    \item \textbf{DDPMs:} Denoising Diffusion Probabilistic Models
    \item \textbf{DeScoD-ECG:} Denoising Score-based Diffusion for Electrocardiogram
    \item \textbf{DiffWave:} Diffusion Waveform
    \item \textbf{DiffDreamer:} Diffusion Dreamer
    \item \textbf{DiffLL:} Diffusion Model for Low-Light
    \item \textbf{DMs:} Diffusion Models
    \item \textbf{DMSEtext:} Diffusion Model for Speech Enhancement text
    \item \textbf{DisC-Diff:} Discriminator Consistency Diffusion
    \item \textbf{DSBID:} Diffusion-based Stochastic Blind Image Deblurring
    \item \textbf{DSC:} Dice Similarity Coefficient
    \item \textbf{DICDNet:} Deep Interpretable Convolutional Dictionary Networks
    \item \textbf{EquiDiff:} Equivariant Diffusion
    \item \textbf{FID:} Frechet Inception Distance
    \item \textbf{GED:} Generalized Energy Distance
    \item \textbf{GNNs:} Graph Neural Networks
    \item \textbf{HiFi-Diff:} Hierarchical Feature Conditional Diffusion
    \item \textbf{HQS:} Hybrid Quality Score
    \item \textbf{ID3PM:} Identity Denoising Diffusion Probabilistic Model
    \item \textbf{IS:} Inception Score
    \item \textbf{KID:} Kernel Inception Distance
    \item \textbf{LEs:} Learnable Unauthorized Examples
    \item \textbf{LDMs:} Latent Diffusion Models
    \item \textbf{LPIPS:} Learned Perceptual Image Patch Similarity
    \item \textbf{MAAT:} Metric Anomaly Anticipation
    \item \textbf{MAD:} Mean Absolute Deviation
    \item \textbf{MAE:} Mean Absolute Error
    \item \textbf{MatFusion:} Material Fusion
    \item \textbf{MOS:} Mean Opinion Score
    \item \textbf{MPJPE:} Mean Per-Joint Position Error
    \item \textbf{NILM:} Non-Intrusive Load Monitoring
    \item \textbf{NASDM:} Nuclei-Aware Semantic Diffusion Model
    \item \textbf{NIQE:} Naturalness Image Quality Evaluator
    \item \textbf{OMOMO:} Object Motion Guided Human Motion Synthesis
    \item \textbf{PatchDDM:} Patch-based Diffusion Denoising Model
    \item \textbf{PRD:} Percent Root Mean Square Difference
    \item \textbf{PSNR:} Peak Signal-to-Noise Ratio
    \item \textbf{RGB-D-Fusion:} Red-Green-Blue Depth Fusion
    \item \textbf{RNNs:} Recurrent Neural Networks
    \item \textbf{RMSE:} Root Mean Square Error
    \item \textbf{SAG:} Self-Attention Guidance
    \item \textbf{SBDMs:} Score-Based Diffusion Models
    \item \textbf{SDEs:} Stochastic Differential Equations
    \item \textbf{SDG:} Semantic Diffusion Guidance
    \item \textbf{SegDiff:} Segmentation Diffusion
    \item \textbf{SketchFFusion:} Sketch-Driven Fusion
    \item \textbf{SMOS:} Style Similarity MOS
    \item \textbf{SSIM:} Structural Similarity Index Measure
    \item \textbf{SDEs:} Stochastic Differential Equations
    \item \textbf{TFDPM:} Temporal and Feature Pattern-based Diffusion Probabilistic Model
    \item \textbf{VDMs:} Variational Diffusion Models
    \item \textbf{VTF-GAN:} Visible-to-Thermal Facial GAN
\end{itemize}
\end{multicols}

\section{Introduction}\label{sec1}

\begin{figure}
    \centering
    \includegraphics{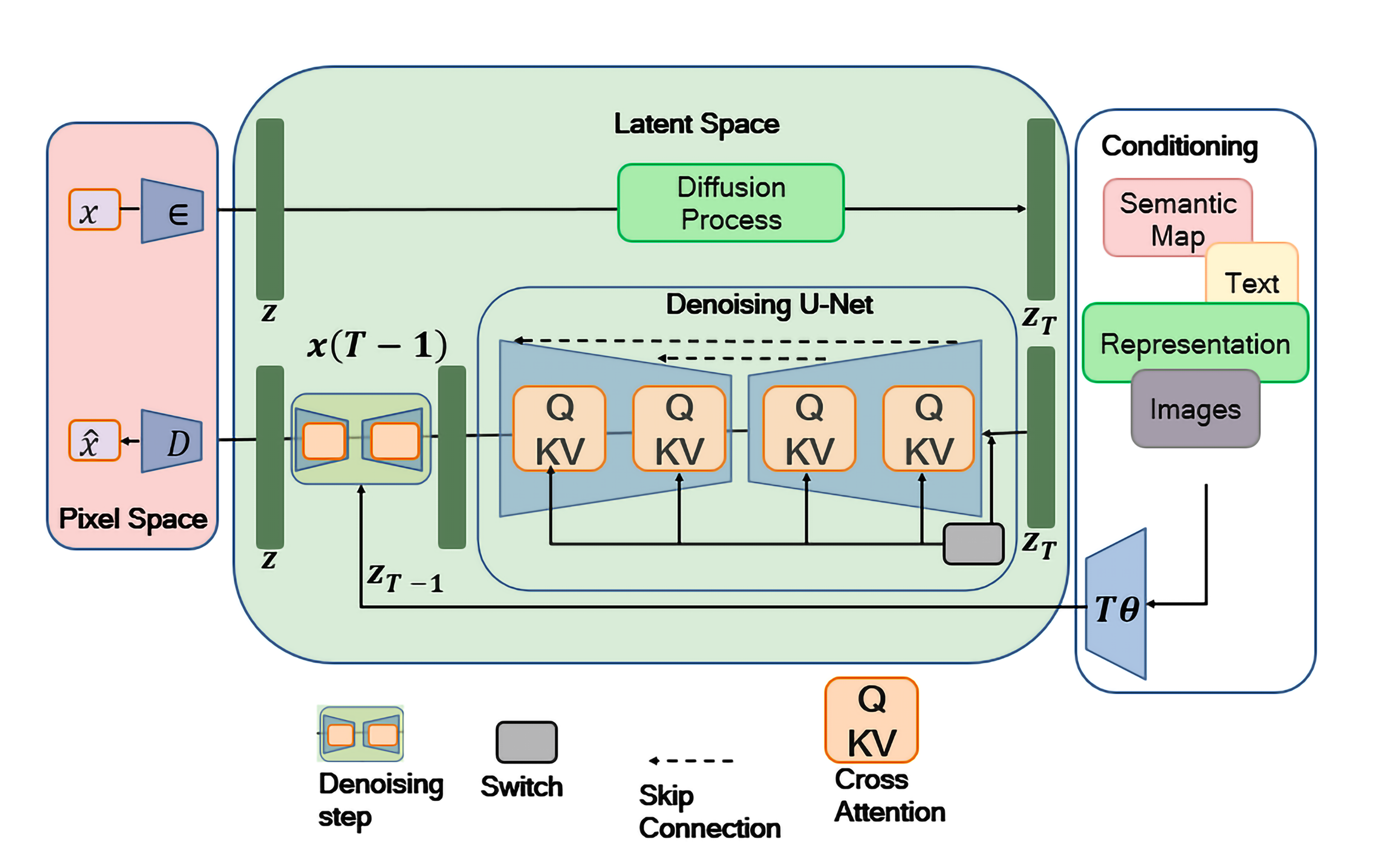}
    \caption{\textbf{An example of Diffusion-based models. From the figure, it can be observed that the model uses cross-attention mechanisms to enhance image synthesis. This approach allows the model to integrate different types of input information, such as text or semantic maps, to control the image generation process more effectively. The figure shows how these inputs are processed and incorporated into the model to produce high-quality images~\cite{rombach2022high}.}}
    \label{fig:difone}
\end{figure}

A \textbf{Diffusion Model} (DM) is a type of generative model that creates data by reversing a diffusion process, which incrementally adds noise to the data until it becomes a Gaussian distribution. First introduced by Sohl-Dickstein et al. (2015), these models have shown exceptional performance in producing high-quality samples across various fields, such as image, audio, and video synthesis~\cite{sohl2015deep,ho2020denoising}. The process involves an iterative procedure where the model is trained to predict the noise that has been added to the sample at each step, effectively learning to denoise data. This approach has led to significant advancements in generating detailed and coherent outputs, making DM a powerful tool for tasks that require high fidelity generation, such as text-to-image synthesis and improving low-resolution images~\cite{saharia2021image}. \textbf{Figure~\ref{fig:difone}} illustrates a DM introduced for high-resolution image synthesis.

Diffusion Models (DMs) have become popular in several areas, particularly in image generation, where they create photorealistic images, art, and edits based on textual descriptions~\cite{saharia2021image, ramesh2021zero}. They are also becoming popular in Natural Language Processing (NLP) for text generation and enhancement, demonstrating an ability to produce coherent and contextually relevant text~\cite{austin2021structured}. In audio synthesis, DMs are used to generate realistic soundscapes, music, and human-like speech, pushing the boundaries of creative and communicative Artificial Intelligence (AI) applications~\cite{kong2020diffwave}. Moreover, their application extends to molecular and material science for designing new chemical compounds and materials, demonstrating their versatility. The popularity of DMs rises from their robustness, flexibility, and the high fidelity of the generated outputs, positioning them as a groundbreaking tool in AI-driven creative and scientific fields~\cite{hoogeboom2022equivariant}.

\begin{figure}[h]
    \centering
    \includegraphics[width=\textwidth]{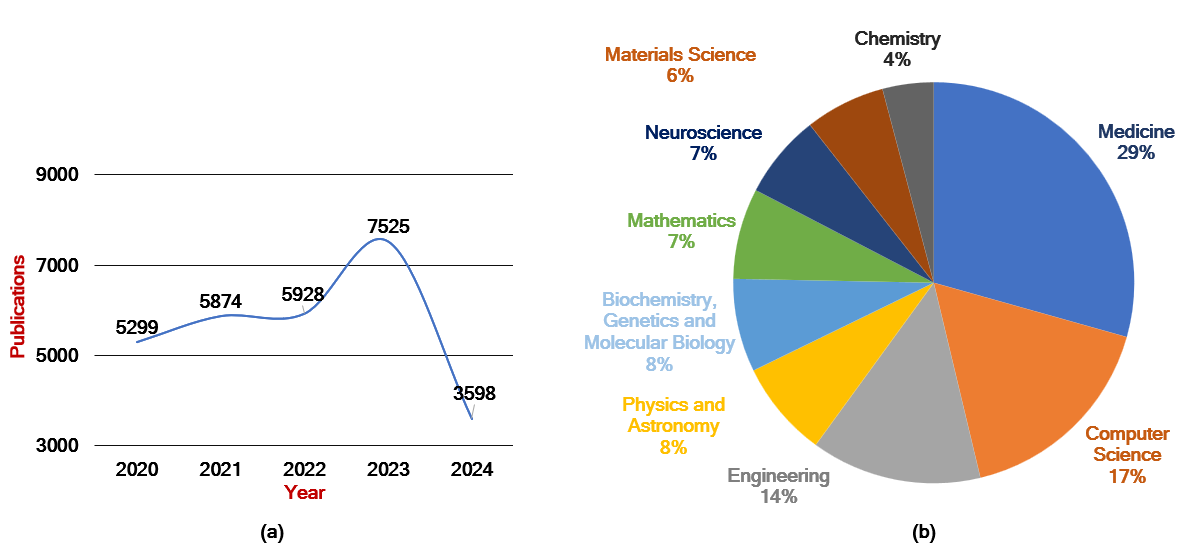}
   \caption{\textbf{Statistics on (a) the number of papers published over the last five years in DMs and (b) the percentage of published papers across various domains.}}
    \label{fig:pubyear}
\end{figure}

\textbf{Figure~\ref{fig:pubyear}} provides a statistical overview of the last five years of published papers on DMs in various disciplines. From \textbf{Figure~\ref{fig:pubyear}(a)}, it can be observed that the number of papers published since 2020 has been constantly growing. \textbf{Figure~\ref{fig:pubyear}(b)} shows that medicine dominates with 29\% of the publications, followed by computer science with 17\% and engineering with 14\%. Fields such as chemistry and materials science have fewer publications, comprising 4\% and 6\% of the total, respectively. These trends highlight the extensive use of DMs in medicine and computer science, while their potential in other areas remains less explored.

This review aims to provide a comprehensive overview of DMs across various domains, helping the general audience understand their ability and versatility. By presenting diverse applications, this review encourages interdisciplinary collaboration and innovation, potentially addressing open challenges in less-explored fields beyond traditional applications like computer vision.

\subsection{Motivation and uniqueness of this survey}
The rapid advancements in DMs across various domains show their potential and versatility. Despite the increasing number of publications, existing surveys often focus on specific applications or narrow fields, leaving a gap in reviews that cover the wide range of DM applications. Considering this opportunity, this survey aims to address the gap in the existing literature by providing a comprehensive overview of DMs. 

Our \textbf{contributions} are summarized below:
\begin{itemize}
    \item[\ding{113}] This survey considers several key aspects of DMs, including theory, algorithms, innovations, media quality, image transformation, healthcare applications, and more. We provide an overview of relevant literature up to March 2024, highlighting the latest techniques and advancements.

    \item[\ding{113}] We categorize DMs into three main types: Denoising Diffusion Probabilistic Models (DDPMs), Noise-Conditioned Score Networks (NCSNs), and Stochastic Differential Equations (SDEs), which aids in understanding their theoretical foundations and algorithmic variations.

    \item[\ding{113}] We highlight novel approaches and experimental methodologies relevant to the application of DMs, considering data types, algorithms, applications, datasets, evaluations, and limitations.

    \item[\ding{113}] Finally, we discuss the findings, identify open issues, and raise questions about future research directions in DMs, aiming to guide researchers and practitioners.
\end{itemize}

\begin{figure}
    \centering
    \includegraphics[width=\textwidth]{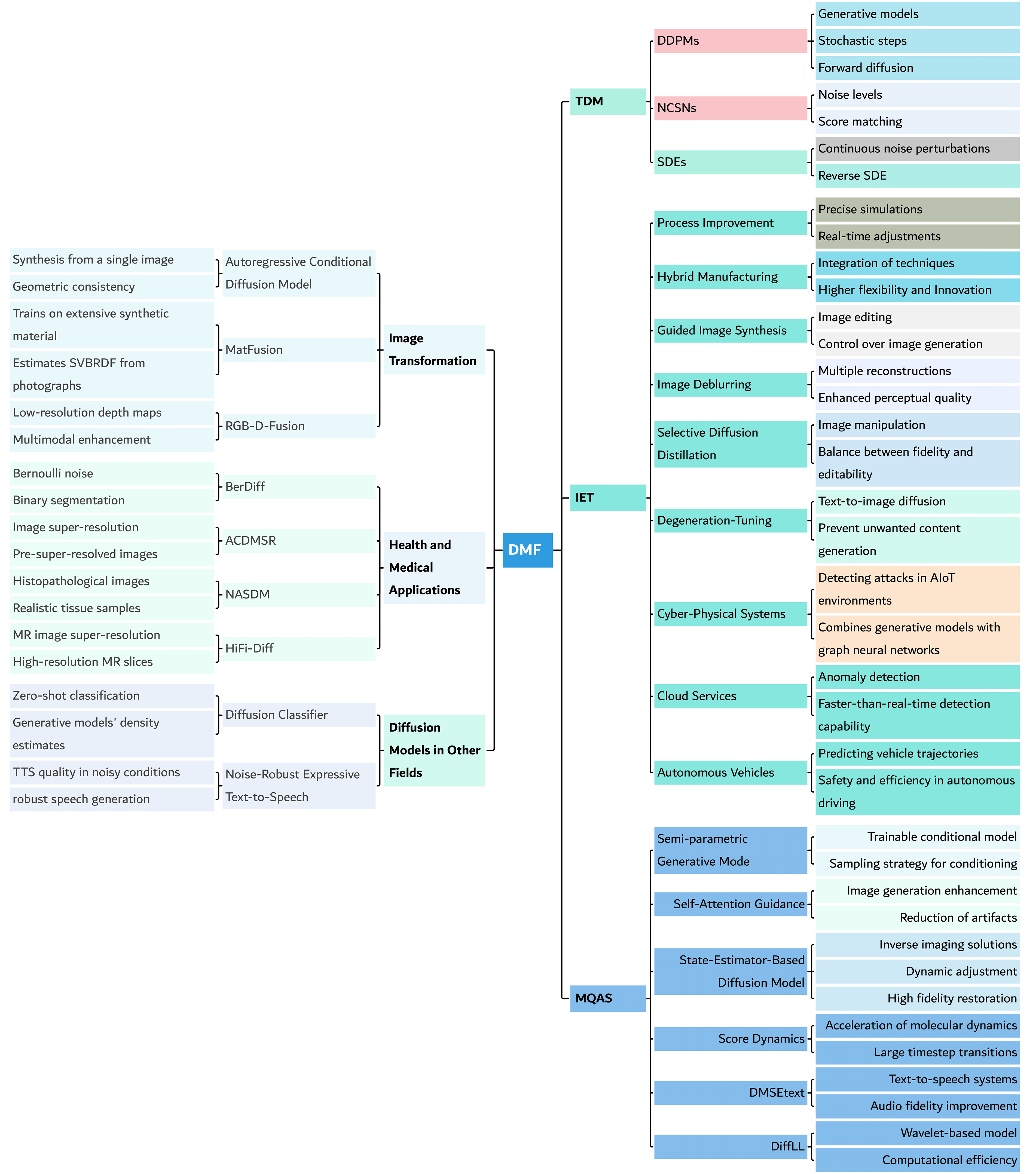}
    \caption{\textbf{Comprehensive overview of DMs: This diagram categorizes various DMs and their applications across different fields. DMF -- Diffusion Models framework, TDM -- Types of Diffusion Models, IET -- Image Enhancement and Transformation, MQAS -- Media Quality, Authenticity, and Synthesis, DDPMs -- Denoising Diffusion Probabilistic Models, NCSNs -- Noise-Conditioned Score Networks, SDEs -- Stochastic Differential Equations.}}
    \label{fig:dttttt}
\end{figure}

\textbf{Figure~\ref{fig:dttttt}} illustrates the framework of DMs based on the referenced literature used in this study, discussed in \textbf{Sections~\ref{dt1}-~\ref{dt6}}.

\subsection{Search strategy}

Data were sourced from Scopus, initially identified 3,746 articles using the title, abstract, and keywords with the search terms `Diffusion Model' AND (`image' OR `audio' OR `text' OR `speech'). Restricting the search to English-language, peer-reviewed, and open-access papers published between 2020 and 2024 reduced the number to 473. Further filtering excluded terms such as `human,' `controlled study,' `job analysis,' `quantitative analysis,' `comparative study,' `specificity,' and other irrelevant keywords, resulting in 326 papers.

One researcher (Y.L.) imported these 326 journal articles into Excel CSV files for detailed analysis. Later, Excel's duplication tools were used to identify and remove duplicates. The titles and abstracts of the remaining papers were assessed by two independent reviewers (M.A. and Z.S.), identifying 65 relevant documents. Additionally, 20 more relevant papers were included, resulting in a total of 85 papers across various fields.

\section{General Overview of DMs}\label{dt1}
DMs are a type of generative model that simulates the diffusion process to construct or reconstruct data distributions through stochastic processes. This involves a dual-phase operation where noise is incrementally added and subsequently reversed~\cite{song2020score}. The algorithmic backbone of DMs contains several key phases~\cite{sohl2015deep, song2020score}:

\begin{itemize}
    \item \textbf{Initialization:} Start with data in its original form \( x_0 \).
    \item \textbf{Forward Process (Noise Addition):} Gradually add noise over \( T \) timesteps, transforming the data from \( x_0 \) to \( x_T \) based on a predefined noise schedule \( \beta \).
    \item \textbf{Reverse Process (Denoising):} Sequentially estimate \( x_{t-1} \) from \( x_t \) using the learned parameters \( \theta \), effectively reversing the noise addition to either reconstruct the original data or generate new data samples.
    
   \item \textbf{Input:} Original data \( X = \{x_1, x_2, \ldots, x_n\} \), Total timesteps \( T \), Noise schedule \( \beta \).

\item \textbf{Output:} Denoised or synthesized data \( X' \).

    \item \textbf{Training:} Train the model to approximate the reverse noise addition process by learning the conditional distributions \( p_\theta(x_{t-1} | x_t) \) for each timestep \( t \), from \( T \) down to 1.
    \item \textbf{Data Synthesis:} Begin with a sample of random noise \( x_T \) and iteratively apply the learned reverse process:
    \[
    x'_{t-1} = \text{Sample\ from\ } p_\theta(x_{t-1} | x_t)
    \]
    culminating in \( x'_0 \), the final synthesized or reconstructed data.

\end{itemize}

\begin{figure}[h]
    \centering
    \includegraphics[width=\textwidth]{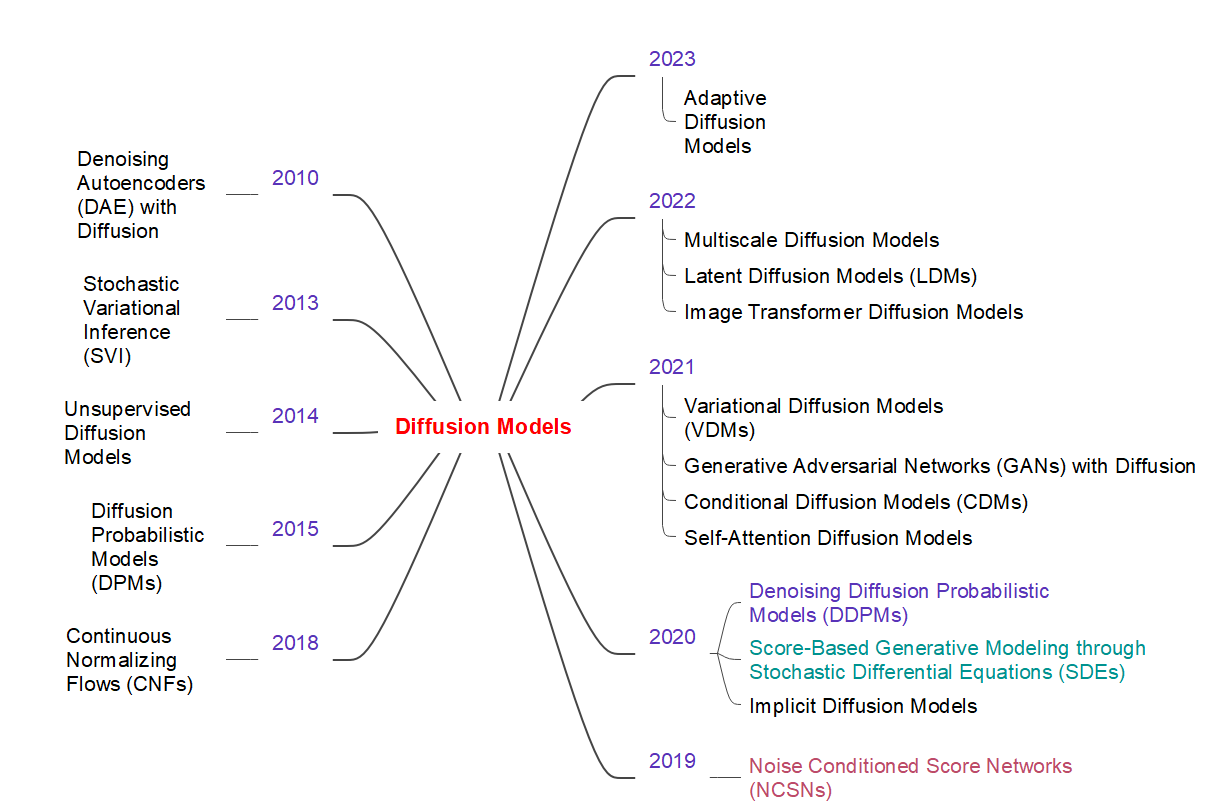}
    \caption{\textbf{Timeline of different DMs from 2010 to 2023. The three main DMs, such as NCSNs, DDPMs, and SDEs, are highlighted with different colors.}}
    \label{fig:timeline}
\end{figure}

\textbf{Types of DM.} Over the years, several diffusion-based models have been proposed, each contributing uniquely to the advancement of generative modeling. \textbf{Figure~\ref{fig:timeline}} illustrates some of the important and influential DM along with their timeline. Among them, three DMs are very popular and widely adopted due to their impact on various applications: DDPMs, NCSNs, and SDEs.

\subsection{DDPMs}\label{ddpms}
Introduced by Ho et al. (2020), DDPMs are generative models that transform noise into data through a series of gradual stochastic steps~\cite{ho2020denoising}.

\textbf{Forward Diffusion Process.} The forward diffusion process incrementally adds Gaussian noise to the data, transforming it into a noise distribution. Given a data point \( \mathbf{x}_0 \), the process is defined over \( T \) timesteps. At each timestep \( t \), Gaussian noise is added to the data:

\begin{equation}
q(\mathbf{x}_t | \mathbf{x}_{t-1}) = \mathcal{N}(\mathbf{x}_t; \sqrt{1 - \beta_t} \mathbf{x}_{t-1}, \beta_t \mathbf{I}),
\end{equation}

By the end of the diffusion process, the data is effectively transformed into pure Gaussian noise.

\textbf{Reverse Denoising Process.} The reverse denoising process aims to recover the original data from the noisy observations. This is modeled using a parameterized reverse Markov chain, where the goal is to estimate the posterior distribution \( q(\mathbf{x}_{t-1} | \mathbf{x}_t) \). However, this posterior is not directly computable, so a Neural Network (NN) \( p_\theta \) is employed to approximate it:

\begin{equation}
p_\theta(\mathbf{x}_{t-1} | \mathbf{x}_t) = \mathcal{N}(\mathbf{x}_{t-1}; \mu_\theta(\mathbf{x}_t, t), \Sigma_\theta(\mathbf{x}_t, t)).
\end{equation}

The network is trained to minimize the following Variational Lower Bound (VLB) on the negative Log-likelihood:

\begin{equation}
L_{\text{vlb}} = \mathbb{E}_{q} \left[ \sum_{t=1}^T D_{\text{KL}} \left( q(\mathbf{x}_{t-1} | \mathbf{x}_t, \mathbf{x}_0) \, || \, p_\theta(\mathbf{x}_{t-1} | \mathbf{x}_t) \right) - \log p_\theta(\mathbf{x}_0 | \mathbf{x}_1) \right],
\end{equation}

\textbf{Training Objective.} To simplify the training process, Ho et al. (2020) proposed a reparameterization of the training objective that aligns closely with denoising score matching. The simplified objective can be expressed as~\cite{ho2020denoising}:

\begin{equation}
L_{\text{simple}} = \mathbb{E}_{t, \mathbf{x}_0, \epsilon} \left[ || \epsilon - \epsilon_\theta(\mathbf{x}_t, t) ||^2 \right],
\end{equation}

\textbf{Sampling from DDPMs.} Once trained, sampling from DDPMs involves running the reverse process starting from pure Gaussian noise \( \mathbf{x}_T \sim \mathcal{N}(0, \mathbf{I}) \) and iteratively applying the learned denoising steps:

\begin{equation}
\mathbf{x}_{t-1} = \mu_\theta(\mathbf{x}_t, t) + \Sigma_\theta(\mathbf{x}_t, t) \cdot \mathbf{z},
\end{equation}
\subsection{NCSNs}

Introduced by Song et al. (2019), NCSNs aim to generate data by estimating the gradients of the data distribution, known as Score Functions, at various noise levels~\cite{song2019generative}.

\textbf{Forward Diffusion Process.} The forward diffusion process in NCSNs involves gradually perturbing the data with Gaussian noise of increasing intensity, similar to the procedure described for \hyperref[ddpms]{\textbf{DDPMs}}. Given an initial data point \( \mathbf{x}_0 \), the data is progressively noised to generate a sequence of noisy data points \( \{\mathbf{x}_t\} \) over \( T \) timesteps.

\textbf{Learning the Score Function.} The core of NCSNs lies in learning the score function, which is the gradient of the log data density \( \nabla_{\mathbf{x}} \log p(\mathbf{x}) \). However, instead of directly learning this for the original data distribution, NCSNs learn it for the perturbed data at various noise levels. An NN \( s_\theta(\mathbf{x}, \sigma_t) \) is trained to approximate these score functions for different noise levels \( \sigma_t \):

\begin{equation}
s_\theta(\mathbf{x}, \sigma_t) \approx \nabla_{\mathbf{x}} \log p_\sigma(\mathbf{x}),
\end{equation}

\textbf{Training Objective.} The training objective for NCSNs involves minimizing a denoising score matching objective, which encourages the NN to accurately predict the score function. This loss function can be expressed as:

\begin{equation}
L_{\text{ncsn}} = \mathbb{E}_{\mathbf{x}_0, \mathbf{\epsilon}, \sigma} \left[ \lambda(\sigma) || s_\theta(\mathbf{x}_0 + \sigma \mathbf{\epsilon}, \sigma) - \frac{\mathbf{\epsilon}}{\sigma} ||^2 \right],
\end{equation}

\textbf{Sampling from NCSNs.} Sampling from NCSNs involves using the learned score function to iteratively denoise a sample of
pure Gaussian noise. This is typically done using Langevin dynamics, a method that iteratively refines the noisy sample by adding the score function and some additional noise:

\begin{equation}
\mathbf{x}_{t+1} = \mathbf{x}_t + \frac{\alpha}{2} s_\theta(\mathbf{x}_t, \sigma_t) + \sqrt{\alpha} \mathbf{z}, \quad \mathbf{z} \sim \mathcal{N}(0, \mathbf{I}),
\end{equation}
\subsection{SDEs}

Introduced by Song et al. (2020), SDEs leverage the mathematical framework of SDE to model the data generation process through continuous noise perturbations and denoising~\cite{song2020score}.

\textbf{Forward Diffusion Process.} In the SDE framework, the forward diffusion process involves transforming the data into a noise distribution through a continuous-time stochastic process, similar to the procedure described for \hyperref[ddpms]{\textbf{DDPMs}}. This is typically modeled by an Itô SDE:

\begin{equation}
d\mathbf{x} = f(\mathbf{x}, t) dt + g(t) d\mathbf{w},
\end{equation}

where \( \mathbf{x} \) represents the data, \( t \) is the time variable, \( f(\mathbf{x}, t) \) is the drift coefficient, \( g(t) \) is the diffusion coefficient, and \( \mathbf{w} \) is a standard Wiener process~\cite{song2020score}.

\textbf{Reverse SDE Process.} The reverse process aims to revert the noisy data back to its original form by solving the reverse-time SDE. This process is governed by:

\begin{equation}
d\mathbf{x} = [f(\mathbf{x}, t) - g(t)^2 \nabla_{\mathbf{x}} \log p_t(\mathbf{x})] dt + g(t) d\mathbf{\bar{w}},
\end{equation}

where \( \nabla_{\mathbf{x}} \log p_t(\mathbf{x}) \) is the score function, which represents the gradient of the log-density of the data at time \( t \), and \( \mathbf{\bar{w}} \) is a standard Wiener process running backward in time~\cite{anderson1982reverse}. The score function is estimated using an NN trained on denoising score matching~\cite{vincent2011connection}.

\textbf{Training Objective.} The training of Score-based Generative models involves learning the score function at different noise levels. The objective function for training is typically the denoising score matching loss:

\begin{equation}
L_{\text{score}} = \mathbb{E}_{p_0(\mathbf{x}), \epsilon, t} \left[ \lambda(t) || s_\theta(\mathbf{x}_t, t) - \nabla_{\mathbf{x}_t} \log p_t(\mathbf{x}_t | \mathbf{x}_0) ||^2 \right],
\end{equation}

\textbf{Sampling from SDEs.} Sampling from the trained Score-based model involves solving the reverse-time SDE starting from a sample of Gaussian noise. Numerical solvers, such as Euler-Maruyama or Predictor-Corrector methods, are used to approximate the reverse SDE and generate data samples~\cite{song2020score}.
\section{General Applications of DMs}

Over the years, interest in DMs has grown exceptionally due to their ability to generate high-quality, realistic, and diverse data samples, making them highly deployable in several cutting-edge applications. Some of the most popular areas where DMs are used extensively include:

\begin{itemize}
    \item[\ding{113}] \textbf{Image Synthesis:} DMs are used to create detailed, high-resolution images from a distribution of noise. They can generate new images or improve existing ones by improving clarity and resolution, making them particularly useful in fields such as digital art and graphic design~\cite{wang2022semantic}.

    \item[\ding{113}] \textbf{Text Generation:} DMs are capable of producing coherent and contextually relevant text sequences. This makes them suitable for applications such as creating literary content, generating realistic dialogues in virtual assistants, and automating content generation for news articles or creative writing~\cite{gong2022diffuseq}.

    \item[\ding{113}] \textbf{Audio Synthesis:} DMs can generate clear and realistic audio from noisy signals. This is valuable in music production, where it's necessary to create new sounds or improve the clarity of recorded audio, as well as in speech synthesis technologies used in various assistive devices~\cite{kong2020diffwave}.

    \item[\ding{113}] \textbf{Healthcare Applications:} Although not limited to medical imaging, DMs assist in synthesizing medical data, including Magnetic Resonance Imaging (MRI), Computed Tomography (CT) scans, and other imaging modalities. This ability is vital for training medical professionals, improving diagnostic tools, and developing more precise therapeutic strategies without compromising patient privacy~\cite{kazerouni2023diffusion}.
\end{itemize}
\textbf{Table~\ref{tab:dif}} summarizes some of the renowned papers in DMs from 2020 to 2023, their proposed algorithms, used datasets, and applications. Different colors are used to distinguish between various algorithms and application types. From \textbf{Table~\ref{tab:dif}}, it can be observed that most of the papers primarily focus on image-based applications, such as image generation, segmentation, and reconstruction.

\begin{longtable}{|p{1cm}|p{3cm}|p{4cm}|p{4cm}|}
\caption{\textbf{Some of the important papers in DMs from 2020 to 2023, along with their proposed algorithms, used datasets, and applications. Different colors are used to distinguish between various algorithms and application types.}}\label{tab:dif}\\
\hline
\rowcolor{gray!20}
\textbf{Year} & \textbf{Proposed Algorithm} & \textbf{Used Datasets} & \textbf{Applications} \\ 
\hline
2020 & \cellcolor{red!20} DDPMs~\cite{ho2020denoising} & CIFAR-10~\cite{krizhevsky2009learning}, LSUN~\cite{yu2015lsun}, CelebA~\cite{liu2015faceattributes} & \cellcolor{blue!20} Image generation \\ 
\hline
2020 & \cellcolor{red!20} Score-Based DMs~\cite{song2020improved} & CIFAR-10, CelebA, LSUN & \cellcolor{blue!20} Image generation \\ 
\hline
2020 & \cellcolor{red!20} SDEs~\cite{song2020score} & CIFAR-10, CelebA, LSUN, FFHQ & \cellcolor{blue!20} Image generation \\ 
\hline
2021 & \cellcolor{red!20} Classifier-guided DMs (CDMs)~\cite{dhariwal2021diffusion} & ImageNet, LSUN, CIFAR-10 & \cellcolor{blue!20} Image generation \\ 
\hline
2021 & \cellcolor{red!20} Variational Diffusion Models (VDMs)~\cite{kingma2021variational} & CIFAR-10, CelebA, LSUN & \cellcolor{blue!20} Image generation \\ 
\hline
2021 & \cellcolor{red!20} Improved DDPMs~\cite{nichol2021improved} & CIFAR-10, CelebA, LSUN & \cellcolor{blue!20} Image generation \\ 
\hline
2021 & \cellcolor{green!20} Diffusion Waveform (DiffWave)~\cite{kong2021diffwave} & LJSpeech, VCTK & \cellcolor{yellow!20} Audio generation \\ 
\hline
2021 & \cellcolor{green!20} Segmentation Diffusion (SegDiff)~\cite{amit2021segdiff} & Cityscapes, Pascal VOC & \cellcolor{cyan!20} Image segmentation \\ 
\hline
2021 & \cellcolor{green!20} Generative LIkelihood-based DEcompression (GLIDE)~\cite{nichol2021glide} & MS-COCO, ImageNet & \cellcolor{lime!20} Image reconstruction \\ 
\hline
2022 & \cellcolor{red!20} Latent Diffusion Models (LDMs)~\cite{rombach2022high} & LAION-400M, CelebA-HQ & \cellcolor{purple!20} Image generation, Text-to-image \\ 
\hline
2022 & \cellcolor{red!20} Image Transformers~\cite{saharia2022image} & ImageNet, COCO & \cellcolor{blue!20} Image generation \\ 
\hline
2022 & \cellcolor{red!20} Multiscale Diffusion Models~\cite{ho2022cascaded} & ImageNet, CIFAR-10, LSUN & \cellcolor{blue!20} Image generation \\  
\hline
2022 & \cellcolor{green!20} Video-DDPM~\cite{ho2022video} & Kinetics-600, UCF-101 & \cellcolor{orange!20} Video generation \\ 
\hline
2023 & \cellcolor{red!20} Adaptive Diffusion Models~\cite{li2023diffusion} & CIFAR-10, CelebA, FFHQ & \cellcolor{blue!20} Image generation \\ 
\hline
\end{longtable}
\section{Innovations and Experimental Techniques in DMs}
\begin{figure}[h]
    \centering
    \includegraphics[width=\textwidth]{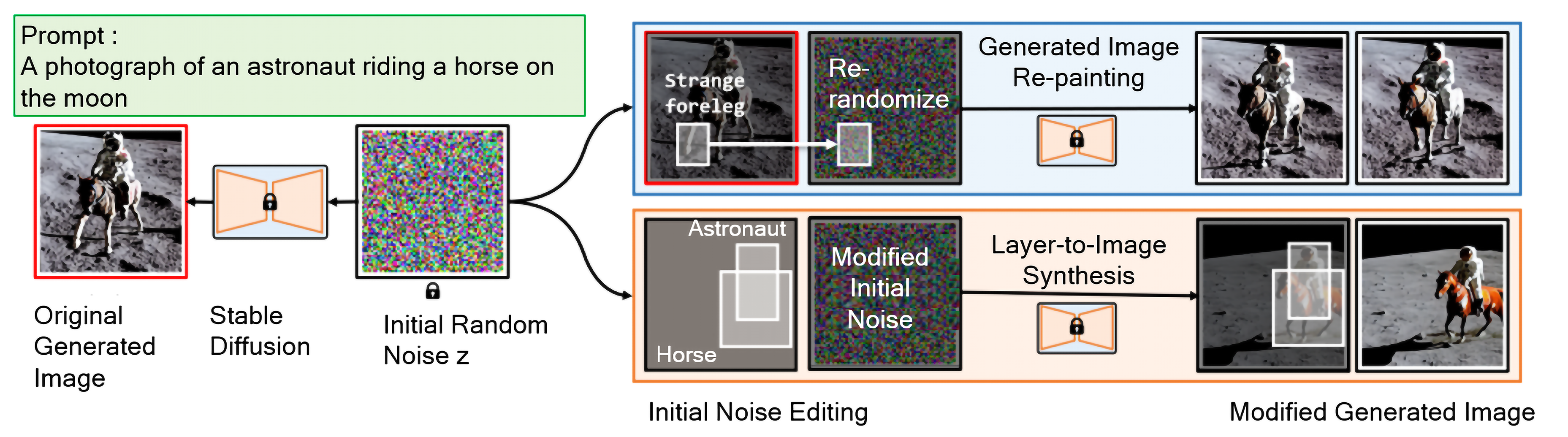}
    \caption{\textbf{Mao et al. (2023) explored how the initial image influenced the image generation process and proposed a new method to control it by altering the initial random noise. They demonstrated two applications: layout-to-image synthesis, which created objects in specified locations, and re-painting, which allowed users to change specific portions of an image while keeping the rest unchanged~\cite{mao2023guided}.}}
    \label{fig:dif2}
\end{figure}

Several studies have utilized DM-based approaches because of their flexibility and effectiveness in various applications. \textbf{Figure~\ref{fig:dif2}} illustrates a DM introduced for guided image synthesis through initial image editing.

Whang et al. (2022) introduced a diffusion-based stochastic blind image deblurring technique. This approach leveraged DMs to produce multiple plausible reconstructions for blurred images, significantly improving perceptual quality. Evaluations on the GoPro dataset showed impressive results with metrics such as FID of 4.04, Kernel Inception Distance (KID) of 0.98, Learned Perceptual Image Patch Similarity (LPIPS) of 0.059, Peak Signal-to-Noise Ratio (PSNR) of 31.66, and Structural Similarity Index Measure (SSIM) of 0.948~\cite{whang2022deblurring}. However, high computational demands pose limitations for real-time applications, suggesting a need for optimized sampling or network architecture adjustments.

Chung et al. (2022) introduced the \textbf{Come-Closer-Diffuse-Faster (CCDF)} sampling strategy to address the slow sampling rate of DMs. CCDF started from a forward-diffused state, reducing required sampling steps using the contraction theory of stochastic difference equations. This method enhanced tasks like super-resolution, image inpainting, and MRI reconstruction, showing improved FID scores and PSNR across datasets~\cite{chung2022come}. However, selecting the optimal starting point remains challenging and requires several trial-and-error approaches.

Wang et al. (2023) introduced \textbf{Selective Diffusion Distillation (SDD)} for improved image manipulation using conditional DMs. SDD trained a feedforward network guided by a DM, addressing the fidelity-editability trade-off. The framework used a Hybrid Quality Score (HQS) to select the optimal semantic timestep, improving image quality and semantic accuracy. SDD outperformed other methods, achieving an FID of 6.066 and a Contrastive Language-Image Pre-training (CLIP) similarity of 0.2337~\cite{wang2023not}. However, a significant limitation remains in the necessity of carefully selecting HQS thresholds to balance manipulation and quality.

Li et al. (2023) introduced \textbf{Object Motion Guided Human Motion Synthesis (OMOMO)}, a framework for synthesizing human motion based on object motion, specifically for large object manipulation. OMOMO used two denoising processes to predict hand positions from object motion and synthesize full-body poses, ensuring accurate contact and realistic motion. By capturing motion via visual-inertial odometry on a smartphone, OMOMO showed potential for applications in virtual reality, augmented reality, and robotics. Their comprehensive dataset demonstrated the framework's ability to generalize to unseen objects. OMOMO achieved high accuracy with a Mean Per-Joint Position Error (MPJPE) of 12.42, a precision score of 0.70, and an F1 score of 0.72~\cite{li2023object}. However, the issue of intermittent object contacts remains unaddressed. Additionally, the predicted hand motions are less plausible, as indicated by lower F1 and precision scores.

Ni et al. (2023) introduced \textbf{Degeneration-Tuning (DT)} to control text-to-image DMs like Stable Diffusion. DT prevents the generation of unwanted content by detaching undesirable textual concepts from image outputs using a scrambled grid. Integrated with Control Network (ControlNet), DT maintains high-quality generation for general content with minimal metric impact (FID from 12.61 to 13.04, IS from 39.20 to 38.25)~\cite{ni2023degeneration}. However, DT's slow sampling speeds, reliance on predefined prompts, and risk of over-degeneration limit its effectiveness, requiring further refinement to balance control and generative abilities.

Yan et al. (2022) introduced \textbf{Temporal and Feature Pattern-based Diffusion Probabilistic Model (TFDPM)}, a model for detecting attacks in cyber-physical systems within Artificial Intelligence of Things (AIoT). TFDPM combined energy-based generative models and Graph Neural Networks to handle complex data and correlations. It extracted temporal and feature patterns to guide a diffusion probabilistic model, improving accuracy and sensitivity~\cite{yan2022tfdpm}. Their proposed TFDPM outperformed many of the existing State-of-the-Art (SOTA) techniques on PUMP, SWAT, and WADI datasets in terms of attack detection accuracy and speed. However, challenges remained in modeling discrete signals and exploring more robust configurations. Additionally, the model faced difficulties in ensuring scalability and adaptability across diverse AIoT environments as well.

Lee et al. (2023) introduced \textbf{Metric Anomaly Anticipation (MAAT)}, a framework for faster-than-real-time anomaly detection in cloud services. MAAT uses a two-stage process: multi-step forecasting with a Conditional Denoising Diffusion Model, followed by anomaly detection with an isolation forest. Tested on AIOps18, Hades, and Yahoo!S5 datasets, MAAT outperformed existing methods in speed, precision, and reliability~\cite{lee2023maat}. However, its focus on cloud-service metrics and a static time horizon limits its applicability to other time-series data and dynamic conditions. Furthermore, its performance with ultra-high-frequency data remains untested, indicating a need for further research to extend its capabilities and validate its effectiveness in these areas.

Chen et al. (2023) introduced \textbf{Equivariant Diffusion (EquiDiff)}, a deep generative model designed to improve the security and efficiency of autonomous vehicles by predicting vehicle routes. EquiDiff uses conditional DMs with an SO(2)-equivariant transformer, integrating historical trajectory data and Gaussian noise to generate future paths while respecting geometric constraints. It also incorporates Recurrent Neural Networks and Graph Attention Networks to model social interactions among vehicles. Evaluated on the NGSIM dataset, EquiDiff outperformed baseline models in short-term prediction accuracy, achieving a Root Mean Square Error (RMSE) of 0.55 at 1 second and 4.01 at 5 seconds~\cite{chen2023equidiff}. However, it showed higher errors in long-term predictions, which highlights limitations in the model's ability to maintain accuracy over extended periods. This suggests the need for further refinement to address these long-term prediction challenges.

\textbf{Table~\ref{tab:mss}} summarizes some of the referenced literature that uses innovative and experimental techniques in developing DMs, including applications in content security, cyber-physical system attack detection, anomaly anticipation, image deblurring, acceleration for inverse problems, image manipulation, and human motion synthesis.

\begin{longtable}{p{.02\linewidth} p{.18\linewidth} p{.18\linewidth} p{.10\linewidth} p{.15\linewidth} p{.20\linewidth}}
\caption{\textbf{Innovations and experimental techniques introduced by the referenced literature in the domain of DMs. FID: Frechet Inception Distance, IS: Inception Score, P\textsubscript{r}: Precision, R\textsubscript{e}: Recall, LPIPS: Learned Perceptual Image Patch Similarity, PSNR: Peak Signal-to-Noise Ratio, SSIM: Structural Similarity Index Measure, KID: Kernel Inception Distance, MPJPE: Mean Per Joint Position Error. Best results are highlighted in bold.}}\label{tab:mss} \\
\toprule
\rowcolor{gray!20} \textbf{Ref.} & \textbf{Algorithms} & \textbf{Applications} & \textbf{Dataset} & \textbf{Evaluations} & \textbf{Limitations} \\
\midrule
\endfirsthead
\caption[]{\textbf{Innovations and experimental techniques introduced by the referenced literature in the domain of DMs (cont.).}} \\
\toprule
\rowcolor{gray!20} \textbf{Ref.} & \textbf{Algorithms} & \textbf{Applications} & \textbf{Dataset} & \textbf{Evaluations} & \textbf{Limitations} \\
\midrule
\endhead
\midrule
\endfoot
\bottomrule
\endlastfoot

\cite{ni2023degeneration} & DT for Content Shielding in Stable DMs & Content shielding in text-to-image Diffusion Models using DT to prevent generation of unwanted concepts & COCO 30K & FID post-DT: 13.04, IS post-DT: 38.25 & DT may limit model's flexibility for diverse contexts. \\
\midrule
\cite{yan2022tfdpm} &  TFDPM & Detecting cyber-physical system attacks using TFDPM with Graph Attention Networks for channel data correlation & PUMP, SWAT, WADI & P\textsubscript{r}: 0.96, R\textsubscript{e}: 0.91, F1: 0.91 & Struggles with discrete signal modeling, needs SDE frameworks for better generative capabilities. \\
\midrule
\cite{lee2023maat} & Maat: Anomaly Anticipation for Cloud Services & Anomaly anticipation using a two-stage Diffusion Model for cloud services, integrating metric forecasting and anomaly detection & AIOps18, Hades, Yahoo!S5 & \textbf{P\textsubscript{r}: 0.97}, \textbf{R\textsubscript{e}: 0.91}, \textbf{F1: 0.91} & Limited generalizability and adaptability post-training. \\
\midrule
\cite{whang2022deblurring} & Diffusion-based Stochastic Blind Image Deblurring & Blind image deblurring using Diffusion Models for multiple reconstructions & GoPro & FID: 4.04, KID: 0.98, LPIPS: 0.06, PSNR: 31.66, SSIM: 0.95 & High computational demands, needs optimized sampling or network architecture. \\
\midrule
\cite{chung2022come} & Come-Closer-Diffuse-Faster & Accelerating CMDs for applications like super-resolution and MRI reconstruction & FFHQ, AFHQ, fastMRI & FID varies; \textbf{PSNR: 33.41} (best MRI case) & Optimal starting values (t0) vary, needs automation for practical deployment. \\
\midrule
\cite{wang2023not} & Selective Diffusion Distillation & Image manipulation balancing fidelity and editability without excessive noise trade-offs & N/A & FID: 6.07, CLIP Similarity: 0.23 & Reliance on correct timestep selection for semantic guidance may limit flexibility. \\
\midrule
\cite{li2023object} & Object Motion Guided Human Motion Synthesis (OMOMO) & Full-body human motion synthesis guided by object motion using a conditional Diffusion Model & Custom dataset & MPJPE: 12.42, Troot: 18.44, Cprec: 0.82, Crec: 0.70, F1: 0.72 & Limited representation of dexterous hand movements and intermittent contact scenarios. \\
\midrule
\cite{ho2020denoising} & DDPMs & Image generation using DDPMs & CIFAR-10, LSUN, CelebA & FID: 3.17, IS: 9.46 & High computational cost and slow sampling speed. \\
\midrule
\cite{song2020improved} & Improved Techniques for Training Score-Based Generative Models & Improved image generation using score-based models & CIFAR-10, CelebA, LSUN & FID: 2.87, IS: 9.68 & Training complexity and large computational resources required. \\
\midrule
\cite{song2020score} & Score-Based Generative Modeling through SDEs & Image generation using SDEs for better quality & CIFAR-10, CelebA, LSUN, FFHQ & FID: 2.92, IS: 9.62 & SDE-based models can be computationally expensive. \\
\midrule
\cite{dhariwal2021diffusion} & Diffusion Models Beat Generative Adversarial Networks (GANs) on Image Synthesis & Image synthesis outperforming GANs using Diffusion Models & ImageNet, LSUN, CIFAR-10 & \textbf{FID: 2.97}, IS: 9.57 & Large model size and slow training times. \\
\midrule
\cite{kingma2021variational} & VDMs & Image generation using variational Diffusion Models & CIFAR-10, CelebA, LSUN & FID: 3.12, IS: 9.53 & Complex model design and high computational cost. \\
\midrule
\cite{nichol2021improved} & Improved Denoising Diffusion Probabilistic Models & Enhanced DDPMs for better image quality & CIFAR-10, CelebA, LSUN & FID: 3.05, IS: 9.50 & Requires extensive hyperparameter tuning. \\
\midrule
\cite{rombach2022high} & High-Resolution Image Synthesis with Latent Diffusion Models (LDMs) & High-resolution image and text-to-image synthesis & LAION-400M, CelebA-HQ & \textbf{FID: 1.97}, IS: 10.32 & High memory usage and computational cost. \\
\midrule
\cite{saharia2022image} & Image Transformers with Autoregressive Models for High-Fidelity Image Synthesis & High-fidelity image synthesis using transformers & ImageNet, COCO & FID: 2.30, IS: 9.95 & Transformer models are computationally intensive. \\
\midrule
\cite{ho2022cascaded} & Cascaded Diffusion Models for High-Fidelity Image Generation & High-fidelity image generation using multiscale Diffusion Models & ImageNet, CIFAR-10, LSUN & FID: 2.15, IS: 9.88 & Cascaded models require extensive computational resources. \\
\midrule
\cite{li2023diffusion} & Optimizing Diffusion Models for Image Synthesis & Adaptive Diffusion Models for better image synthesis & CIFAR-10, CelebA, FFHQ & \textbf{FID: 1.89}, IS: 10.45 & Adaptive models can be complex and resource-intensive. \\
\midrule
\cite{kong2021diffwave} &  DiffWave & Audio generation using Diffusion Models & LJSpeech, VCTK & FID: 3.67, PSNR: 34.10 & High computational cost and slow sampling speed. \\
\midrule
\cite{ho2022video} & Video Diffusion Models (Video-DDPM) & Video generation using Diffusion Models & Kinetics-600, UCF-101 & FID: 3.85, SSIM: 0.92 & High computational demands and slow training times. \\
\midrule
\cite{amit2021segdiff} & SegDiff & Image segmentation using Diffusion Models & Cityscapes, Pascal VOC & FID: 3.50, SSIM: 0.87 & Limited scalability to larger datasets. \\
\midrule
\cite{nichol2021glide} & GLIDE & Photorealistic image generation and editing with text guidance & MS-COCO, ImageNet & FID: 3.21, IS: 9.67 & Text-guided models require extensive training data. \\
\end{longtable}

\section{Media Quality, Authenticity, and Synthesis}\label{dt3}
\begin{figure}[h]
    \centering
    \includegraphics[width=\textwidth]{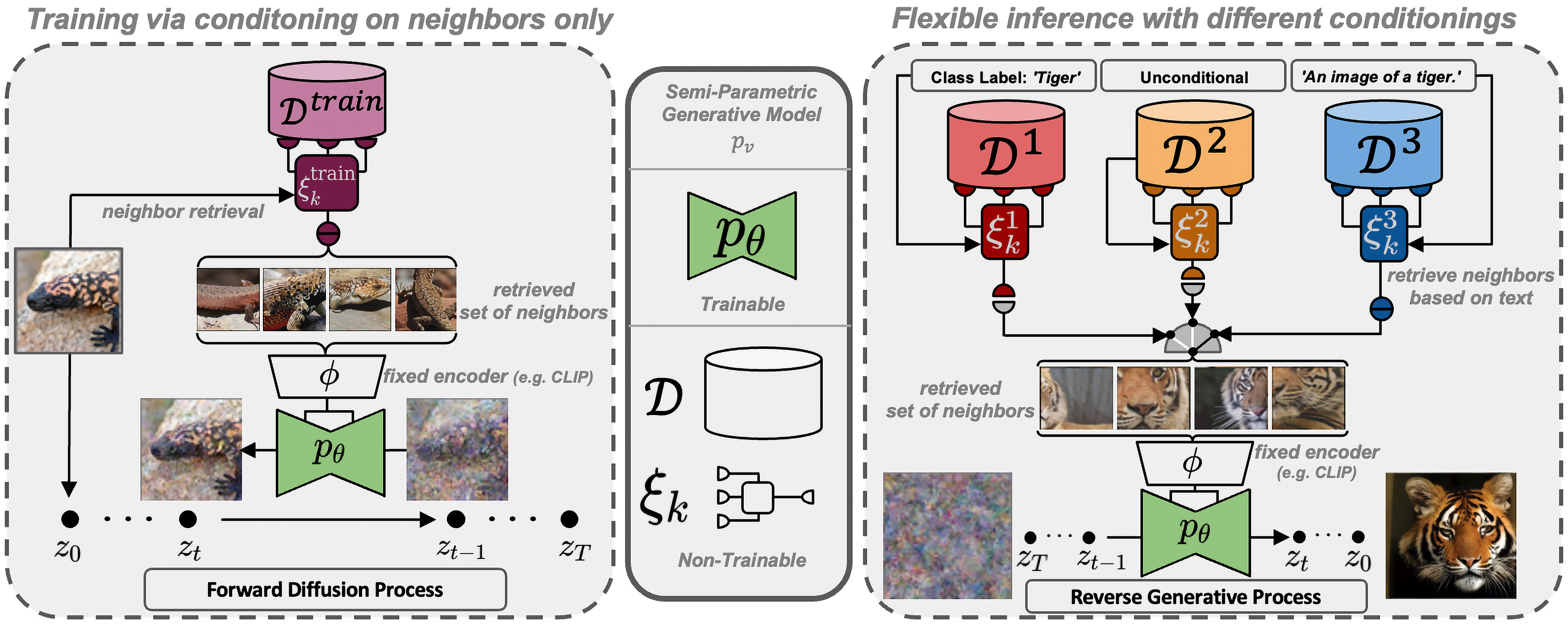}
    \caption{\textbf{A semi-parametric generative model consists of a trainable Conditional Generative Model \( p_\theta(x|\cdot) \), an external database \( D \) with visual examples, and a sampling strategy \( \xi_k \) that selects a subset \( M(k)_D \subseteq D \) for conditioning \( p_\theta \). To train \( p_\theta \) to create consistent scenes using \( M(k)_D \), \( \xi_k \) retrieves the nearest neighbors of each target example from \( D \). By adjusting \( D \) and \( \xi_k \) during inference, the model can flexibly sample with post-hoc conditioning on class labels (\( \xi^1_k \)) or text prompts (\( \xi^3_k \)), and perform zero-shot stylization~\cite{blattmann2022retrieval}.}
}
    \label{fig:mediaf}
\end{figure}

Several studies propose DMs to improve media quality and create realistic samples. \textbf{Figure~\ref{fig:mediaf}} illustrates an orthogonal, semi-parametric DM, which includes a trainable Conditional Generative Model, an external database for visual examples, and a sampling strategy to retrieve subsets for conditioning the model~\cite{blattmann2022retrieval}.

Hong et al. (2023) introduced \textbf{Self-Attention Guidance (SAG)} to improve image generation using Denoising Diffusion Models (DDMs). SAG leverages self-attention maps to focus on significant areas, reducing artifacts and improving image quality. Evaluation on various platforms revealed that SAG significantly improved both FID and IS compared to existing methods~\cite{hong2023improving}. 

Ji et al. (2024) introduced a \textbf{Learnable State-Estimator-based DMs} for inverse imaging problems, restoring clean images from corrupted inputs with high fidelity. This method uses a state estimator to dynamically adjust the diffusion process within a latent space, achieving computational efficiency and avoiding extensive training. Evaluated on tasks like inpainting, deblurring, and JPEG compression restoration, it showed strong performance, particularly on the FFHQ dataset with a PSNR of 27.98, LPIPS of 0.0939, and FID of 25.453~\cite{ji2024diffusion}. However, the model relies on current generative abilities and needs domain-specific adaptations for broader applications.

Tian et al. (2023) introduced \textbf{Diffusion Model for Speech Enhancement text (DMSEtext)}, a conditional DMs designed to enhance speech quality in Text-to-Speech (TTS) systems by addressing audio degradations. Operating in the log Mel-spectrogram domain, it uses text transcriptions to improve audio fidelity. DMSEtext achieved a Mean Opinion Score (MOS) of 4.32 for cleanliness and 4.17 overall, with a reduced Phoneme Error Rate (PER) of 17.6\%, indicating improved clarity and authenticity~\cite{tian2023diffusion}. However, its performance depends on the quality of text transcription and varies under different audio types.

Jiang et al. (2023) introduced \textbf{Diffusion Model for Low-Light (DiffLL)}, a framework for improving low-light images using a Wavelet-based Conditional Diffusion Model. This model increases inference speed and reduces computational demands while maintaining high image quality. A High-Frequency Restoration Module improves image details. DiffLL outperformed current methods on benchmarks like LOL-v1, LOLv2-real, and LSRW in PSNR, SSIM, LPIPS, and FID metrics~\cite{jiang2023low}. However, it struggles with extremely low-light conditions and is not optimized for real-time video processing. Additionally, the study did not consider real-time video support and handling diverse lighting conditions, which remain areas for further investigation.

Dong et al. (2023) proposed \textbf{Controlled Language-Image Pretraining Sonic (CLIPSonic)}, a text-to-audio synthesis method using unlabeled videos and pretrained language-vision models. It employs a conditional DMs to generate audio by translating text embeddings into image embeddings, improving zero-shot modality transfer. CLIPSonic demonstrated competitive performance on VGGSound and MUSIC datasets~\cite{dong2023clipsonic}. However, its effectiveness is limited by the quality of pretrained models, distribution mismatches, and training complexity, posing scalability challenges.

Liu et al. (2023) proposed \textbf{Semantic Diffusion Guidance (SDG)}, a framework that improves DDMs with fine-grained control using language, image, or both modalities. SDG integrates guidance into pretrained models via image-text or image matching score gradients which eliminates the need for retraining. It enables text-guided image synthesis on datasets without text annotations using CLIP-based guidance and demonstrates better accuracy over baseline models such as Iterative Latent Variable Refinement~\cite{choi2021ilvr} and StyleGAN+CLIP. On the FFHQ dataset, the proposed SDG models achieved an FID score of 14.37 and a top-1\% accuracy of 0.520. Additionally, the ablation studies on LSUN showed minor performance improvements with different scaling factors~\cite{liu2023more}. However, SDG's effectiveness depends on the accuracy of pretrained models and their ability to process guidance signals. Additionally, the framework poses potential risks of misuse which necessitates the requirement of ethical guidelines to ensure responsible deployment.

Cai et al. (2023) introduced \textbf{Diffusion Dreamer (DiffDreamer)}, an unsupervised framework for scene extrapolation using conditional DMs to generate novel views from given images. By training on internet-collected nature images, DiffDreamer refines projected RGBD images through guided denoising steps, conditioned on multiple past and future frames. It significantly outperforms previous GAN-based methods in quality and consistency. On the LHQ dataset, DiffDreamer achieved an FID score of 51.0 over 100 steps and 34.49 over 20 steps~\cite{cai2023diffdreamer}. However, DiffDreamer cannot synthesize novel views in real time due to the computational intensity of DMs and does not ensure content diversity in extended extrapolations.

Carrillo et al. (2023) proposed an interactive approach for line art colorization using conditional Diffusion Probabilistic Models, allowing users to input initial color strokes. The system integrates these inputs via a dual conditioning strategy, producing diverse, high-quality images. Their model outperforms SOTA methods by achieving an SSIM of 0.81, LPIPS of 0.14, and FID of 6.15. However, the model's accuracy depends on the quality of user input, and the complex conditioning strategy may cause computational inefficiencies, which could eventually affect scalability as well~\cite{carrillo2023diffusart}.

Mao et al. (2023) introduced \textbf{Sketch-Driven Fusion (SketchFFusion)}, a model for sketch-guided image editing using a conditional Diffusion Model. SketchFFusion maintains the integrity of sketches while editing, simulating human sketch styles and preserving structural details. On the CelebA-HQ dataset, it outperformed SOTA methods with an FID of 9.07, PSNR of 26.74, and SSIM of 0.8822. The model was also tested on the COCO-AIGC dataset, demonstrating adaptability across various scenes and objects~\cite{mao2023sketchffusion}. However, SketchFFusion currently only supports binary sketches, limiting its use to black-and-white inputs.

Luo et al. (2023) introduced \textbf{Semantic-Conditional Diffusion Networks} for image captioning, leveraging DMs to improve visual-language alignment and coherence. Unlike traditional transformer models, their approach uses semantic priors from cross-modal retrieval and refines captions through multiple Diffusion Transformer layers. This dynamic integration of image and text features enhances caption relevance and accuracy. On the Common Objects in Context dataset, it achieved a Consensus-based Image Description Evaluation score of 131.6 and a BLEU-4 score of 39.4, outperforming SOTA models~\cite{luo2023semantic}.

\textbf{Table~\ref{tab:medau}} summarizes some of the referenced literature that proposes different diffusion-based approaches to improve media quality and increase authenticity.
\begin{small}
\begin{longtable}{p{.02\linewidth} p{.18\linewidth} p{.18\linewidth} p{.10\linewidth} p{.20\linewidth} p{.15\linewidth}}
\caption{\textbf{Improving media quality using diffusion-based approaches as demonstrated in the existing literature. FID: Frechet Inception Distance, IS: Inception Score, P\textsubscript{r}: Precision, R\textsubscript{e}: Recall, LPIPS: Perceptual Image Patch Similarity, PSNR: Peak Signal-to-Noise Ratio, SSIM: Structural Similarity Index Measure, KID: Kernel Inception Distance, MPJPE: Mean Per Joint Position Error. Best results are highlighted in bold.}}\label{tab:medau} \\
\toprule
\rowcolor{gray!20} \textbf{Ref.} & \textbf{Algorithms} & \textbf{Applications} & \textbf{Dataset} & \textbf{Evaluations} & \textbf{Limitations} \\
\midrule
\endfirsthead
\caption[]{\textbf{Improving media quality using diffusion-based approaches as demonstrated in the existing literature (cont.).}} \\
\toprule
\rowcolor{gray!20} \textbf{Ref.} & \textbf{Algorithms} & \textbf{Applications} & \textbf{Dataset} & \textbf{Evaluations} & \textbf{Limitations} \\
\midrule
\endhead
\midrule
\endfoot
\bottomrule
\endlastfoot

\cite{hong2023improving} & SAG in DDMs & Image generation improvement & ImageNet, LSUN & FID: \textbf{2.58}, sFID: 4.35 & Needs broader application integration. \\
\midrule
\cite{ji2024diffusion} & Learnable State-Estimator-Based Diffusion Model & Inverse imaging problems (inpainting, deblurring, JPEG restoration) & FFHQ, LSUN-Bedroom & PSNR: 27.98, LPIPS: 0.09, FID: 25.45 & Limited generative capabilities, needs domain adaptation. \\
\midrule
\cite{hsu2023score} & Score Dynamics (SD) & Accelerating molecular dynamics simulations & Alanine dipeptide, short alkanes in aqueous solution & Wall-clock speedup up to 180X & Requires large datasets; generalization challenges. \\
\midrule
\cite{tian2023diffusion} & CMDs for Speech Enhancement (DMSEtext) & Speech enhancement for TTS model training & Real-world recordings & MOS Cleanliness: \textbf{4.32} ± 0.08, Overall Impression: \textbf{4.17} ± 0.06, PER: 17.6\% & Needs text conditions for best results. \\
\midrule
\cite{yan2023towards} & Conditional Diffusion Model for HDR Reconstruction & HDR image reconstruction from LDR images & Benchmark datasets for HDR imaging & PSNR-µ: 44.11, PSNR-L: 41.73, SSIM-µ: \textbf{0.99}, SSIM-L: \textbf{0.99}, HDR-VDP-2: 65.52, LPIPS: 0.01, FID: 6.20 & Slow inference speed; improve distortion metrics. \\
\midrule
\cite{dong2023clipsonic} & CLIPSonic & Text-to-audio synthesis using unlabeled videos & VGGSound, MUSIC & FAD: CLIPSonic-ZS on MUSIC 19.30, CLIPSonic-PD on MUSIC 13.51; CLAP score: CLIPSonic-ZS on MUSIC 0.28, CLIPSonic-PD on MUSIC 0.25 & Performance drop in zero-shot modality transfer. \\
\midrule
\cite{liu2023more} & SDG & Fine-grained image synthesis with text and image guidance & FFHQ, LSUN & FID: 14.37 (image guidance on FFHQ), 28.38 (text guidance on FFHQ); Top-5 Retrieval Accuracy: 0.742 (image guidance), 0.878 (text guidance) & Potential misuse in image generation. \\
\midrule
\cite{cai2023diffdreamer} & DiffDreamer: Conditional Diffusion Model for Scene Extrapolation & Unsupervised 3D scene extrapolation & LHQ, ACID & Achieves low FID scores across various step intervals, e.g., 20 steps: FID: 34.49; 100 steps: FID: 51.00 on LHQ & Real-time synthesis not feasible; limited content diversity. \\
\midrule
\cite{carrillo2023diffusart} & Diffusart: Conditional Diffusion Probabilistic Models for Line Art Colorization & Interactive line art colorization with user guidance & Danbooru2021 & SSIM: 0.81, LPIPS: 0.14, FID: \textbf{6.15} & Bias towards white; limits color diversity. \\
\midrule
\cite{mao2023sketchffusion} & SketchFFusion: A Conditional Diffusion Model for Sketch-guided Image Editing & Sketch-guided image editing for local fine-tuning using generated sketches & CelebA-HQ, COCO-AIGC & FID: 9.07, PSNR: 26.74, SSIM: 0.88 & Supports only binary sketches; limits color editing. \\
\midrule
\cite{luo2023semantic} & Semantic-Conditional Diffusion Networks for Image Captioning & Advanced text-to-image captioning using semantic-driven Diffusion Models & COCO & B@1: 79.0, B@2: 63.4, B@3: 49.1, B@4: 37.3, CIDEr: \textbf{131.6} & Lacks real-time processing; needs optimization. \\
\midrule
\cite{chen2023equidiff} & EquiDiff: Deep Generative Model for Vehicle Trajectory Prediction & Trajectory prediction for autonomous vehicles using a deep generative model with SO(2)-equivariant transformer & NGSIM & RMSE for 5s trajectory prediction shows competitive results & Effective short-term; higher errors in long-term predictions. \\
\midrule
\cite{peng2023generating} & Efficient MRI Synthesis with Conditional Diffusion Probabilistic Models & Efficient synthesis of 3D brain MRIs using a conditional Diffusion Model & ADNI-1, UCSF, SRI International & MS-SSIM: 78.6\% & Focused on T1-weighted MRIs; explore more types. \\
\end{longtable}
\end{small}

\section{Image Transformation and Enhancement}\label{dt4}
\subsection{Image-to-image transformation}
DMs have shown significant potential in various image-to-image transformation tasks. Existing studies demonstrate that the versatility of DMs helps in improving image quality and generating new images. For instance, Yu et al. (2023) presented an autoregressive \textbf{Cascade Multiscale Diffusion (CMD) for Novel View Synthesis (NVS)} from a single image, ensuring photorealistic and geometrically consistent image sequences. They introduced the Thresholded Symmetric Epipolar Distance for evaluating geometric consistency. Their proposed model outperforms GeoGPT and LookOut models when tested on CLEVR, RealEstate10K, and Matterport3D datasets in terms of LPIPS and PSNR. For example, on RealEstate10K, it achieves an LPIPS of 0.333 and a PSNR of 15.51 compared to LookOut's 0.378 and 14.43~\cite{yu2023long}. However, the model faces limitations, including performance drops in certain conditions and challenging scenarios. For instance, it may struggle with images that have complex textures or dynamic elements, leading to less accurate geometric consistency and lower visual quality. Additionally, the model's robustness in diverse real-world environments is not fully tested, indicating a need for further refinement and evaluation to ensure reliable performance across a wider range of situations.

Yin et al. (2023) introduced \textbf{Controllable Light Enhancement (CLE)} Diffusion, a novel framework for low-light image enhancement that offers users dynamic control over brightness adjustments. Utilizing CMDs with an illumination embedding and integrating the Segment-Anything Model (SAM), CLE Diffusion allows precise, region-specific improvements. Their proposed approach outperformed existing models in terms of PSNR, SSIM, LPIPS, and LI-LPIPS on the LOL and MIT-Adobe FiveK datasets. For instance, on the LOL dataset, it achieved a PSNR of 25.51 and an SSIM of 0.89~\cite{yin2023cle}. However, the slow inference speeds hinder real-time application and usability in time-sensitive scenarios. Additionally, the model struggles to maintain high performance in environments with complex and varying lighting conditions, leading to inaccuracies and lower image quality in such settings.

\begin{figure}[h]
    \centering
    \includegraphics[width=\textwidth]{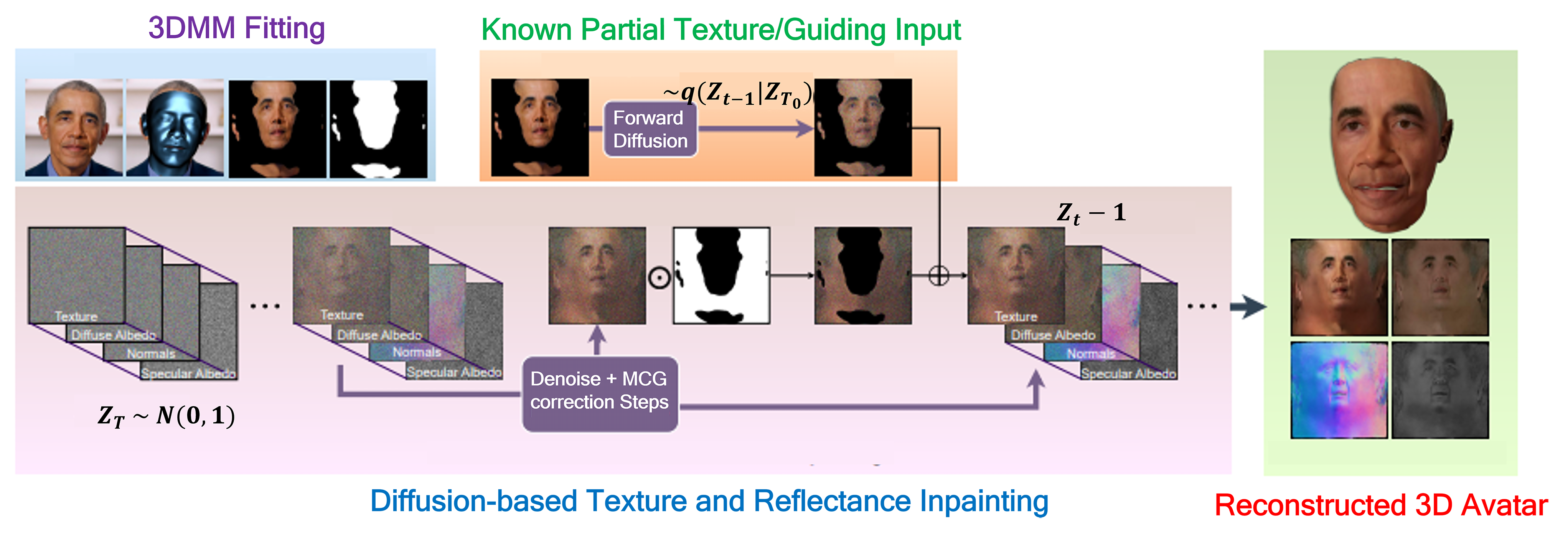}
    \caption{\textbf{The Relightify method employs a latent Diffusion Model for inference, visualizing denoising in the original image space. It initiates with 3DMM fitting to generate a partial UV texture via image-to-UV rasterization. The process then uses random noise, guided by known texture, to complete missing pixels in the texture/reflectance diffusion model. Denoising steps (\(z_t\) to \(z_{t-1}\), \(t \in \{1, \ldots, T\}\)) follow an inpainting approach similar to MCG: 1) Updating reflectance maps and unobserved texture pixels using reverse diffusion sampling and manifold constraints, and 2) Directly sampling known pixels from the input texture through forward diffusion (\(\odot\) and \(\oplus\) denote the Hadamard product and addition). Masking is applied solely to the texture, while reflectance maps (diffuse/specular albedo, normals) are predicted entirely from random noise. This technique produces high-quality rendering assets for realistic 3D avatar creation~\cite{papantoniou2023relightify}.}}
    \label{fig:papu}
\end{figure}

Papantoniou et al. (2023) introduced \textbf{``Relightify,"} a method for 3D facial \textbf{Bidirectional Reflectance Distribution Function (BRDF)} reconstruction from a single image using DMs (\textbf{Figure~\ref{fig:papu}}). Relightify is trained on a UV dataset of facial reflectance to understand facial features and lighting interactions. It fits a 3D model to an input image, unwraps the face into a UV texture, and uses the Diffusion Model to fill in occluded areas while keeping the original textures for realistic results. Relightify outperforms methods like CE, UV-GAN, and OSTeC, especially in handling different viewing angles, as measured by its higher PSNR and SSIM metrics~\cite{papantoniou2023relightify}.

Kirch et al. (2023) presented \textbf{Red-Green-Blue Depth Fusion (RGB-D-Fusion)}, a multi-modal conditional diffusion denoising model that enhanced depth map resolution from low-resolution RGB images of humanoid subjects. Unlike Variational Autoencoders or GANs, RGB-D-Fusion employed diffusion denoising models in two stages: creating and refining low-resolution depth maps with RGB-D images, incorporating depth noise augmentation for robustness. It effectively generated detailed depth maps represented as point clouds when tested on a dataset of 25k samples~\cite{kirch2023rgb}. However, it required substantial resources for sampling and training and relied on known projection matrices, limiting its scalability and flexibility.

Mao et al. (2023) improved \textbf{multi-contrast MRI using the Discriminator Consistency Diffusion (DisC-Diff)} model, which stabilizes and leverages multi-contrast data. DisC-Diff outperforms existing techniques in terms of PSNR and SSIM metrics when tested on normal and pathological brain datasets~\cite{mao2023disc}. Nonetheless, the study has limitations, such as the risk of mode collapse when processing multi-contrast MRI data, which can impact the reliability of the super-resolution process. Additionally, the proposed DMs may not adequately capture the complex interactions in multi-contrast MRI, limiting their effectiveness in clinical applications.

\textbf{Table~\ref{tab:imtim}} summarizes some of the referenced literature that proposes different diffusion-based approaches for image-to-image transformation.

\begin{small}
\begin{longtable}{p{.02\linewidth} p{.13\linewidth} p{.12\linewidth} p{.14\linewidth} p{.20\linewidth} p{.20\linewidth}}
\caption{\textbf{Image-to-image transformation using different DMs. FID: Frechet Inception Distance, LPIPS: Learned Perceptual Image Patch Similarity, PSNR: Peak Signal-to-Noise Ratio, SSIM: Structural Similarity Index Measure, MSE: Mean Squared Error, RMSE: Root Mean Squared Error, CD: Chamfer Distance, EMD: Earth Mover's Distance, IoU: Intersection over Union, VLB: Visible Light Blocking. Best results are highlighted in bold.}}\label{tab:imtim} \\
\toprule
\rowcolor{gray!20} \textbf{Ref.} & \textbf{Algorithms} & \textbf{Applications} & \textbf{Dataset} & \textbf{Evaluations} & \textbf{Limitations} \\
\midrule
\endfirsthead
\caption[]{\textbf{Image-to-image transformation using different DMs (cont.).}} \\
\toprule
\rowcolor{gray!20} \textbf{Ref.} & \textbf{Algorithms} & \textbf{Applications} & \textbf{Dataset} & \textbf{Evaluations} & \textbf{Limitations} \\
\midrule
\endhead
\midrule
\endfoot
\bottomrule
\endlastfoot

\cite{yu2023long} & Autoregressive conditional Diffusion-based Models (ACDM) & NVS from a single image & RealEstate10K, MP3D, CLEVR & LPIPS: 0.33, PSNR: 15.51 on RealEstate10K; LPIPS: 0.50, PSNR: 14.83 on MP3D; FID: 26.76 on RealEstate10K; FID: 73.16 on MP3D & Requires complex geometric consistency and heavy computational resources for extrapolating views. \\ \midrule
\cite{yin2023cle} & CLE Diffusion & Low light enhancement & LOL, MIT-Adobe FiveK & PSNR: 29.81, SSIM: \textbf{0.97} on MIT-Adobe FiveK; PSNR: 25.51, SSIM: 0.89, LPIPS: 0.16, LI-LPIPS: 0.18 on LOL & Slow inference speed and limited capability in handling complex lighting and blurry scenes. \\ \midrule
\cite{papantoniou2023relightify} & Diffusion-based inpainting model for 3D facial BRDF reconstruction & Facial texture completion and reflectance reconstruction from a single image & MultiPIE & PSNR: 26.00, SSIM: 0.93 at 0° angle on MultiPIE; Sampling time: 17 sec & Limited by input image quality and potential under-representation of ethnic diversity in training data. \\ \midrule
\cite{yan2023towards} & CMDs & HDR reconstruction from multi-exposed LDR images & Benchmark datasets for HDR imaging & PSNR-µ: 22.25, SSIM-µ: 0.84, LPIPS: 0.03 on Hu's dataset & Slow inference speed due to iterative denoising process. \\ \midrule

\cite{kirch2023rgb} & RGB-D-Fusion diffusion probabilistic models & Depth map generation and super-resolution from monocular images & Custom dataset with $\approx$ 25,000 RGB-D images from 3D models of people & MSE: 1.48, IoU: \textbf{0.99}, VLB: 16.95 with UNet3+ model & High computational resources required for training and sampling. \\ \midrule
\cite{mao2023disc} & Disentangled CMDs (DisC-Diff) & Multi-contrast MRI super-resolution & IXI dataset and clinical brain MRI dataset & PSNR: \textbf{37.77} dB, SSIM: \textbf{0.99} on 2× scale in clinical dataset & Requires accurate condition sampling for model precision. \\ \hline

\end{longtable}
\end{small}

\subsection{Image quality enhancement and processing}

DMs have been used effectively in various image quality improvement tasks, as shown in \textbf{Table~\ref{tab:imh}}. These tasks include improving document images, generating thermal facial images, and creating identity-preserving face images. In each case, DMs significantly boost image quality. They make images clearer, remove noise and watermarks, and create realistic images in different conditions, showing their versatility in image processing~\cite{yang2023docdiff, ordun2023visible, kansy2023controllable, yu2023freedom}.

Yang et al. (2023) introduced \textbf{Document Diffusion (DocDiff)}, a diffusion-based framework for restoring degraded document images. This framework recovers low-frequency content using a Coarse Predictor and high-frequency details with a High-Frequency Residual Refinement (HRR) module. DocDiff's efficient architecture achieves SOTA results on benchmarks, improving readability and text edge sharpness with only 4.17 million parameters in the HRR module~\cite{yang2023docdiff}. However, the study did not consider additional document improvement tasks such as document super-resolution or style transfer. The robustness of DocDiff in handling various types and levels of document degradation might be insufficient, requiring more diverse training data and improved network architectures. Furthermore, there is no user-centric evaluation to assess the impact on readability and user satisfaction.

Ordun et al. (2023) introduced \textbf{Visible-to-Thermal Facial GAN (VTF-GAN)}, a generative adversarial network that created high-resolution thermal facial images from visible spectrum inputs, addressing the lack of thermal sensors in common RGB cameras for telemedicine~\cite{ordun2023visible}. However, the study did not address the potential biases or ethical considerations that may arise when using generated thermal faces for applications such as telemedicine, which could be crucial in real-world implementations. There is a lack of analysis regarding the generalizability of the VTF-GAN model across different datasets or demographic groups, which could impact its applicability in diverse scenarios. Additionally, there is no comprehensive discussion on the interpretability of the generated images and how they align with the underlying physiological conditions they aim to represent.

Kansy et al. (2023) introduced the Identity \textbf{Denoising Diffusion Probabilistic Model (ID3PM)}, which can reverse-engineer face recognition models without needing full access (i.e., using a black-box method) to the model. ID3PM uses denoising diffusion to generate high-quality, identity-preserving facial images without needing an identity-specific loss. It effectively samples from the inverse distribution, producing diverse images with varying backgrounds, lighting, poses, and expressions~\cite{kansy2023controllable}. Nonetheless, the method presented in this work generates images at a relatively low resolution of 64 x 64, which may limit the fine details captured. Additionally, the inference times for image generation are relatively long, and small artifacts in the output could affect the overall quality of the generated images.

\begin{figure}
    \centering
    \includegraphics[width=\textwidth]{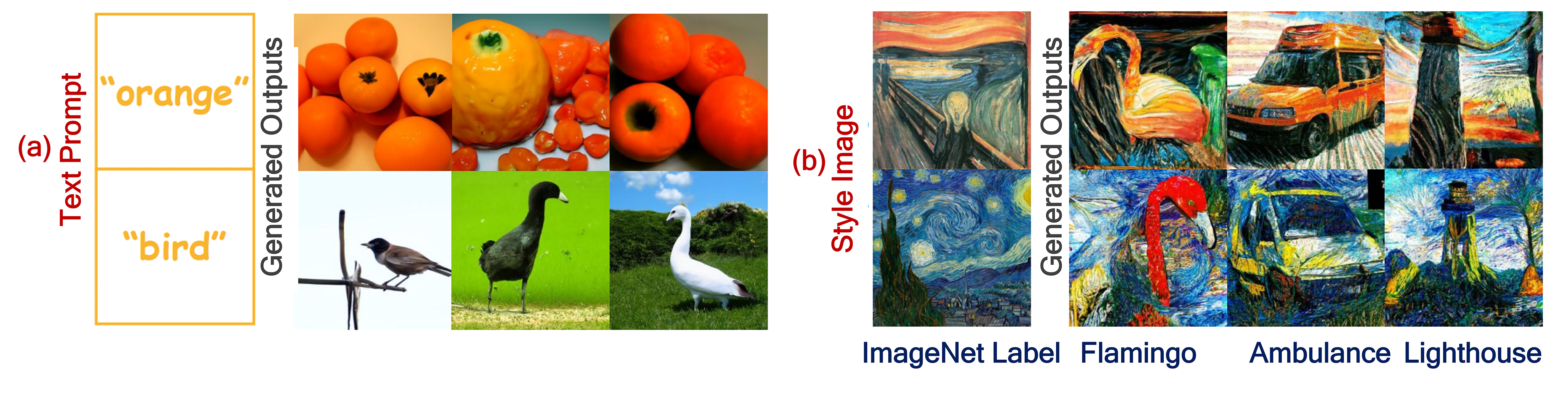}
    \caption{\textbf{Single-condition guided results based on FreeDoM models, where (a) are Unconditional DMs and (b) are Classifier-based DMs, generated output on the ImageNet dataset~\cite{yu2023freedom}.}}
    \label{fig:freedom}
\end{figure}

Yu et al. (2023) presented \textbf{Free-form Deformation Model (FreeDoM)}, a versatile training-free conditional Diffusion Model that adapts to various conditions without condition-specific training (\textbf{Figure~\ref{fig:freedom}}). Unlike traditional models, FreeDoM uses pre-trained networks to create time-independent energy functions, reducing costs and improving transferability. Tested on different data domains, it outperforms training-required methods like Text- and Image-driven Generative Adversarial Network (TediGAN) in generating segmentation maps, sketches, and text-conditioned images, with better condition matching and FID scores~\cite{yu2023freedom}. While FreeDoM is designed to be training-free and adaptable to various conditions, it may struggle in situations where the conditions significantly differ from the capabilities of the pre-trained networks. This limitation could affect the model's effectiveness in diverse and complex scenarios.

\begin{small}
\begin{longtable}{p{.02\linewidth} p{.13\linewidth} p{.13\linewidth} p{.11\linewidth} p{.16\linewidth} p{.22\linewidth} p{.22\linewidth}}
\caption{\textbf{Image Enhancement and Processing based on the referenced literature. FID: Frechet Inception Distance, LPIPS: Learned Perceptual Image Patch Similarity, PSNR: Peak Signal-to-Noise Ratio, SSIM: Structural Similarity Index Measure, MANIQA: Mean Opinion Score Quality Index, MUSIQ: Measurement Uncertainty Simulation Quality Index, DISTS: Deep Image Structure and Texture Similarity, MSE: Mean Squared Error. Best results are highlighted in bold.}}\label{tab:imh} \\
\toprule
\rowcolor{gray!20} \textbf{Ref.} & \textbf{Algorithms} & \textbf{Applications} & \textbf{Dataset} & \textbf{Evaluations} & \textbf{Limitations} \\
\midrule
\endfirsthead
\caption[]{\textbf{Image Enhancement and Processing (cont.).}} \\
\toprule
\rowcolor{gray!20} \textbf{Ref.} & \textbf{Algorithms} & \textbf{Applications} & \textbf{Dataset} & \textbf{Evaluations} & \textbf{Limitations} \\
\midrule
\endhead
\midrule
\endfoot
\bottomrule
\endlastfoot

\cite{yang2023docdiff} & DocDiff conditional Diffusion Model & Document image enhancement including deblurring, denoising, and watermark removal & Document Deblurring Dataset & MANIQA: 0.72, MUSIQ: 50.62, DISTS: 0.06, LPIPS: 0.03, PSNR: 23.28, SSIM: \textbf{0.95} & May lose high-frequency information, leading to distorted text edges. Relies on the quality of low-frequency content recovery by the Coarse Predictor module. \\ \midrule

\cite{ordun2023visible} & VTF-GAN & Thermal facial imagery generation for telemedicine & Eurecom and Devcom datasets & FID: 47.35, DBCNN: 34.34\%, MSE: 0.88, SPEC: -1.1\% for VTF-GAN with Fourier Transform-Guided (FFT-G) & Generation constrained to static environments; performance untested in dynamic, variable conditions affecting thermal emission. \\ \midrule

\cite{kansy2023controllable} & ID3PM & Inversion of pre-trained face recognition models, generating identity-preserving face images & LFW, AgeDB-30, CFP-FP datasets & LFW: \textbf{99.20}\%, AgeDB-30: 94.53\%, CFP-FP: 96.13\% with ID3PM using InsightFace embeddings & Generation quality may vary with the diversity of embeddings; control over the generation process might need fine-tuning for specific applications. \\ \midrule

\cite{yu2023freedom} & FreeDoM & Conditional image and latent code generation & Multiple datasets for segmentation maps, sketches, texts & Distance: 1696.1, FID: 53.08 for segmentation maps with FreeDoM & High sampling time cost; struggles with fine-grained control in large data domains; may produce poor results with conflicting conditions. \\ \hline

\end{longtable}
\end{small}

\section{Healthcare and medical applications}\label{dt5}
DMs have made significant contributions to the field of healthcare and medical analysis by offering cutting-edge solutions for a variety of tasks. Models like PatchDDM, a memory-efficient patch-based DM, have been effectively utilized for applications such as tumor segmentation in medical imaging datasets like BraTS2020, showing their ability to generate precise three-dimensional segmentations~\cite{bieder2023diffusion}. Furthermore, DMs are renowned for their extensive mode coverage and the quality of samples they generate. These models are employed in medical imaging to address challenges related to limited data availability, inconsistent data acquisition methods, and privacy issues. For example, the Med-DDPM, a DM-based approach, has demonstrated super stability and performance in comparison to GANs when it comes to generating high quality, realistic 3D medical images~\cite{amirhossein2022diffusion,bieder2023diffusion}. 

Chen et al. (2023) introduced the \textbf{Bernoulli Diffusion Model (BerDiff)} for medical image segmentation. BerDiff used Bernoulli noise instead of Gaussian noise, improving binary segmentation tasks essential in medical imaging. By sampling Bernoulli noise and intermediate latent variables, BerDiff generated diverse and accurate segmentation masks. This approach, tested on the LIDC-IDRI and BRATS 2021 datasets, outperformed SOTA methods in metrics such as Generalized Energy Distance (GED) and Dice score~\cite{chen2023berdiff}. However, the proposed BerDiff model mainly focused on binary image segmentation, which may limit its application to more complex segmentation scenarios such as multi-class tasks. The study did not extensively discuss whether additional post-processing steps were needed for specific clinical tasks.

\begin{figure}
    \centering
    \includegraphics[width=\textwidth]{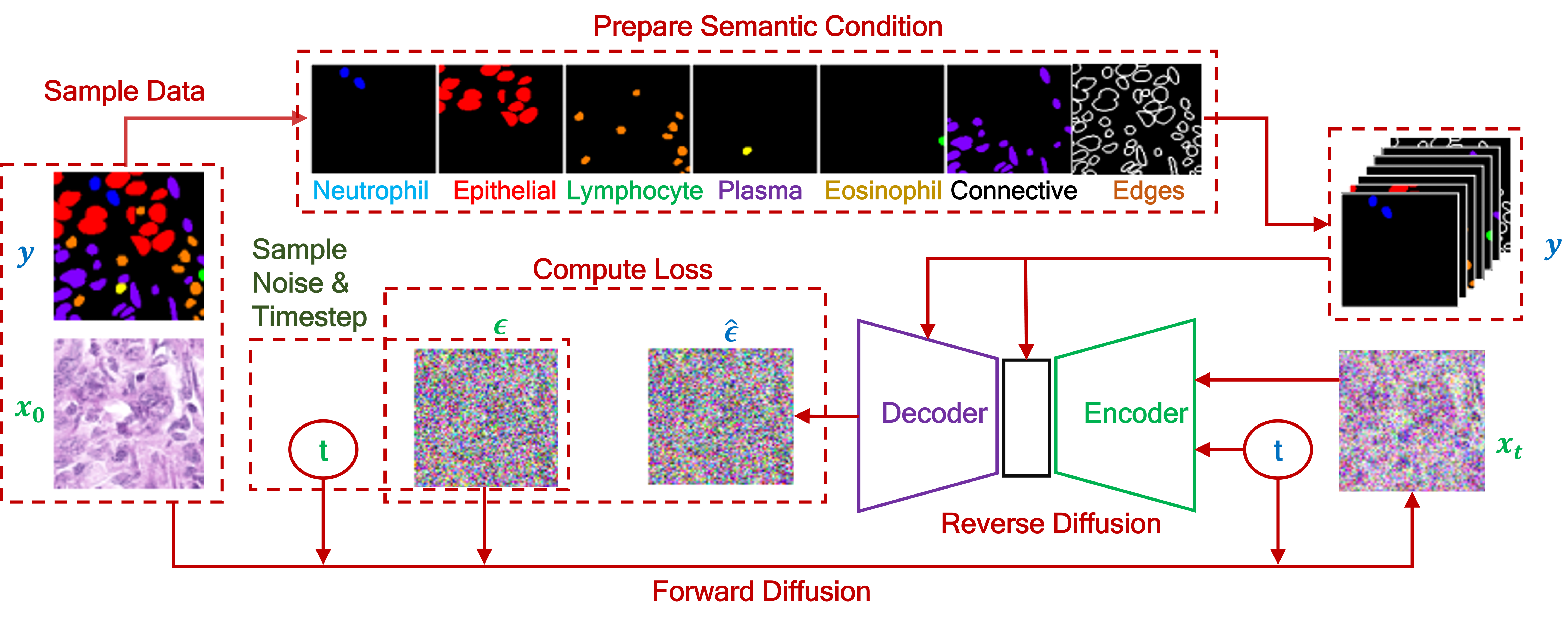}
    \caption{\textbf{The NASDM training protocol initiates with an original image \(x_0\) and its corresponding semantic mask \(y\). It then generates a conditioning signal by enhancing the mask and incorporating an adjacent edge map. Subsequently, a timestep \(t\) is selected, and noise is applied to \(x_0\) through forward diffusion, resulting in a perturbed input \(x_t\). The denoising model then processes this corrupted image \(x_t\), along with the timestep \(t\) and semantic condition \(y\), to estimate \(\hat{\epsilon}\), representing the model's prediction of the total noise introduced. The loss is then computed by comparing this estimate \(\hat{\epsilon}\) with the actual noise \(\epsilon\) applied during the forward diffusion process~\cite{shrivastava2023nasdm}}}
    \label{fig:medonetwo}
\end{figure}

Shrivastava et al. (2023) presented \textbf{Nuclei-Aware Semantic Diffusion Model (NASDM)}, a framework for generating high-quality histopathological images using conditional Diffusion Modeling (\textbf{Figure~\ref{fig:medonetwo}}). NASDM creates realistic tissue samples from semantic instance masks of six nuclei types, aiding pathological analysis and addressing training data scarcity for nuclei segmentation. On a colon dataset, NASDM achieved an FID of 15.7 and an IS of 2.7, outperforming existing methods~\cite{shrivastava2023nasdm}. However, the proposed approaches required large amounts of annotated data for training Deep Learning models for nuclei segmentation, which can be expensive and time-consuming. Additionally, the current methods focused only on generating tissue patches conditioned on the semantic layouts of nuclei, which may have restricted the framework's scope to specific types of histopathological images.

Wang et al. (2023) proposed a novel model, \textbf{Hierarchical Feature Conditional Diffusion (HiFi-Diff)}, a framework for MRI image super-resolution that adapts to varying inter-slice spacings in clinical settings. HiFi-Diff uses hierarchical feature extraction to iteratively convert Gaussian noise into high-resolution MR slices, achieving superior image quality. Tested on the HCP-1200 dataset, HiFi-Diff outperformed traditional methods in PSNR, SSIM, and Dice similarity coefficient across various scaling tasks (×4, ×5, ×6, ×7). For instance, in a ×4 task, it achieved a PSNR of 39.50 and an SSIM of 0.98~\cite{wang2023arbitrary}. While the experimental results demonstrate the effectiveness of HiFi-Diff on the HCP-1200 dataset, the study did not provide any insights regarding the model's performance compared to existing super-resolution methods on a wider range of MRI datasets with varying characteristics.

Li et al. (2023) introduced \textbf{Denoising Score-based Diffusion for Electrocardiogram (DeScoD-ECG)}, a conditional Score-based Diffusion Model for improving Electrocardiogram (ECG) signals, which are essential for diagnosing cardiovascular diseases but often suffer from noise. Unlike traditional Deep Learning methods, DeScoD-ECG iteratively reconstructs signals from Gaussian white noise using a Markov Chain, improving reconstruction quality with a multi-shot averaging strategy. Validated on the QT Database and MIT-BIH Noise Stress Test Database, DeScoD-ECG outperforms existing methods in metrics such as Sum of Squared Differences (SSD), Mean Absolute Deviation (MAD), Percent Root Mean Square Difference (PRD), and Cosine Similarity, showing over a 20\% improvement~\cite{li2023descod}. However, the study did not address other types of noise interference that can affect ECG signals, such as muscle artifacts or electrode motion artifacts. While the study highlights the potential of the DeScoD-ECG model for biomedical applications, it does not discuss any specific real-world applications or case studies where the method has been successfully applied.

\textbf{Table~\ref{tab:hlm}} summarizes some of the existing reference literature that considers DM-based approaches for developing realistic samples in medical imaging and healthcare.
\begin{small}
\begin{longtable}{p{.03\linewidth} p{.15\linewidth} p{.16\linewidth} p{.15\linewidth} p{.16\linewidth} p{.17\linewidth}}
\caption{\textbf{Health and medical applications using diffusion-based approaches as demonstrated in the existing literature. FID: Frechet Inception Distance, IS: Inception Score, PSNR: Peak Signal-to-Noise Ratio, SSIM: Structural Similarity Index Measure, GED: Generalized Energy Distance, HM-IoU: Harmonic Mean Intersection over Union, LPIS: Learned Perceptual Image Similarity. Best results are highlighted in bold.}}\label{tab:hlm} \\
\toprule
\rowcolor{gray!20} \textbf{Ref.} & \textbf{Algorithms} & \textbf{Applications} & \textbf{Dataset} & \textbf{Evaluations} & \textbf{Limitations} \\
\midrule
\endfirsthead
\caption[]{\textbf{Health and medical applications (cont.).}} \\
\toprule
\rowcolor{gray!20} \textbf{Ref.} & \textbf{Algorithms} & \textbf{Applications} & \textbf{Dataset} & \textbf{Evaluations} & \textbf{Limitations} \\
\midrule
\endhead
\midrule
\endfoot
\bottomrule
\endlastfoot

\cite{chen2023berdiff} & BerDiff: Conditional Bernoulli Diffusion for Medical Image Segmentation & Advanced medical image segmentation using Bernoulli diffusion to produce accurate and diverse segmentation masks & LIDC-IDRI, BRATS 2021 & Achieves state-of-the-art performance with metrics on LIDC-IDRI - GED: 0.24, HM-IoU: 0.60, and on BRATS 2021 - Dice: \textbf{89.7}. & Focuses only on binary segmentation and requires significant time for iterative sampling. \\ \midrule

\cite{shrivastava2023nasdm} & NASDM: Nuclei-Aware Semantic Tissue Generation Framework & Generative modeling of histopathological images conditioned on semantic instance masks & Colon dataset & FID: 15.7, IS: 2.7, indicating high-quality and semantically accurate synthetic image generation. & Further development is required for varied histopathological settings and end-to-end tissue generation that includes mask synthesis. \\ \midrule

\cite{wang2023arbitrary} & Hierarchical Feature Conditional Diffusion (HiFi-Diff) & MR image super-resolution with arbitrary reduction of inter-slice spacing & HCP-1200 dataset & PSNR: 39.50±2.29, SSIM: \textbf{0.99} for ×4 SR task & Slow sampling speed, suggesting potential improvements through faster algorithms or knowledge distillation. \\ \hline

\end{longtable}
\end{small}

\section{Applications of Diffusion Models in Other Fields}\label{dt6}

DMs are adopted in various domains beyond image analysis and are effectively used for time series forecasting, imputation, and generation, demonstrating their versatility in handling sequential data. Additionally, DMs have been adapted for predicting chaotic dynamical systems, offering uncertainty quantification and the ability to represent outliers and extreme events effectively. Furthermore, recent advancements have extended DMs to Riemannian manifolds, enabling applications in constrained conformational modeling of protein backbones and robotic arms, highlighting their relevance in scientific domains as well. The evolution of DMs beyond image analysis underscores their adaptability and effectiveness across a wide range of fields~\cite{finzi2023user,yang2024directional,li2023comparison}.

\begin{figure}
    \centering
    \includegraphics[width=\textwidth]{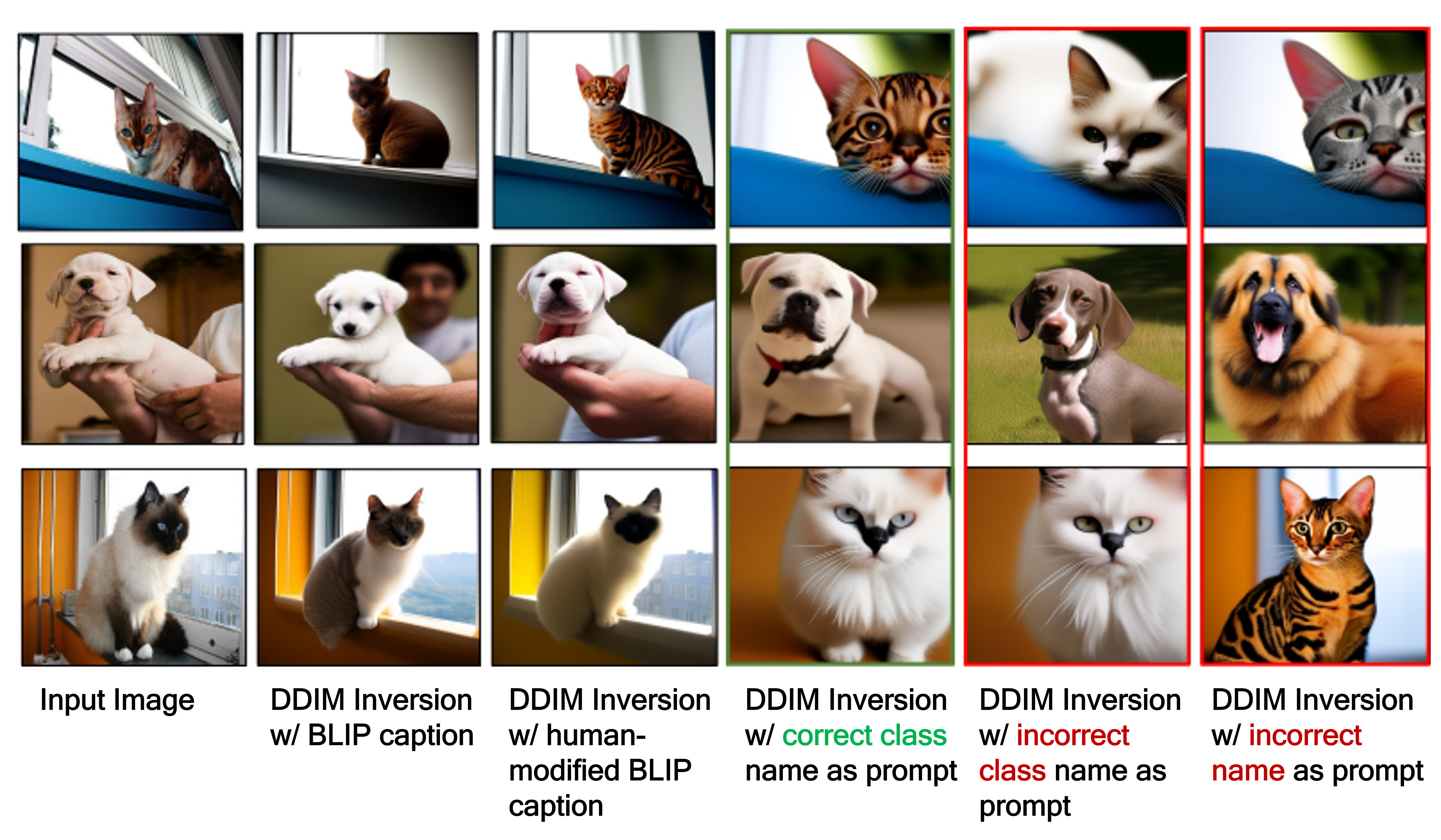}
    \caption{\textbf{Various texts and captions (BLIP, Human-modified BLIP, correct class names, incorrect class names) in zero-shot classification using text-based DMs is examined. The input image is inverted with the caption and reconstructed using deterministic DDIM sampling. Human-modified BLIP captions align best with the input image. Images reconstructed with correct class names (col. 4) match better than those with incorrect class names (col. 5 and 6). In Row 3 (col. 4 and 5), the base Stable DMs fails to distinguish between Birman and Ragdoll breeds, causing classifier failure. BLIP: Bootstrapped Language-Image Pretraining, DDIM: Denoising Diffusion Implicit Models~\cite{li2023your}}}
    \label{fig:your}
\end{figure}

Li et al. (2023) developed the \textbf{Diffusion Classifier}, a novel method using large-scale text-to-image Diffusion Models for zero-shot classification (\textbf{Figure~\ref{fig:your}}). This approach leverages Diffusion Models' density estimates to classify images without additional training, outperforming existing methods. The Diffusion Classifier performs exceptionally well in benchmarks and multimodal compositional reasoning, showing notable improvements in zero-shot reasoning tasks. It also demonstrated robustness against distribution shifts when tested with ImageNet~\cite{li2023your}. While the study focuses on Stable Diffusion, it does not explore the potential challenges or limitations that may arise when applying this approach to other types of classification problems beyond image data.
Zhuang et al. (2023) explored DMs for semantic image synthesis, focusing on abdominal CT images. They compared three models—Conditional DDPM, Mask-guided DDPM, and Edge-guided DDPM—against SOTA GAN-based approaches. By using semantic masks to guide synthesis, the proposed approaches surpassed GANs in terms of FID, PSNR, SSIM, and Dice Score, generating higher-quality and more clinically accurate images~\cite{zhuang2023semantic}. Despite their advantages, the proposed DMs faced significant challenges due to high computational costs and long processing times.

\begin{figure}
    \centering
    \includegraphics[width=\textwidth]{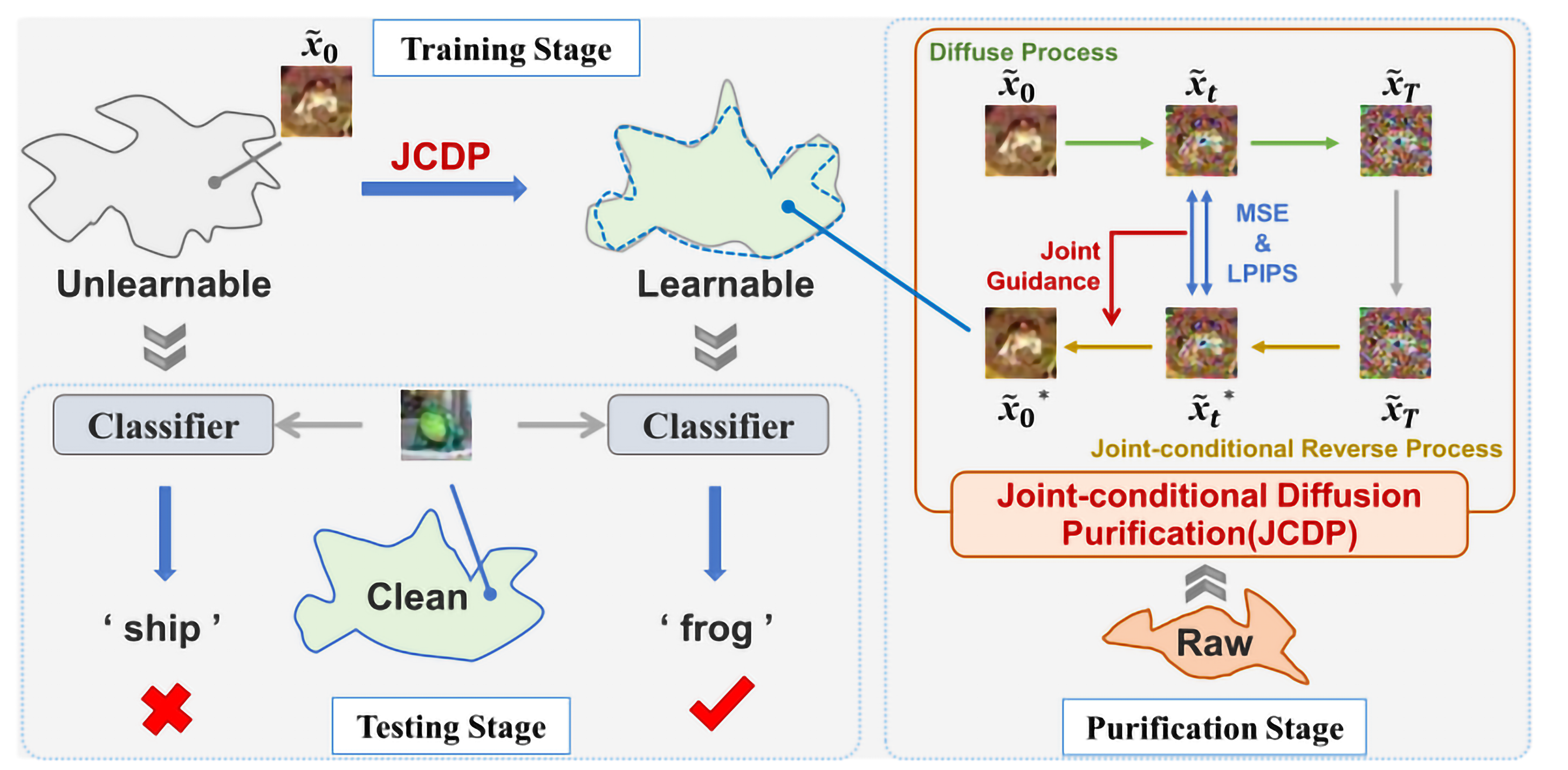}
    \caption{\textbf{Joint-conditional diffusion purification (JCDP) demonstrates the concept of learnable examples. When applied to datasets, non-generalizable or unlearnable data points fail to achieve effective generalization, consequently impacting the quality and reliability of the samples employed in training classification models~\cite{jiang2023unlearnable}.}}
    \label{fig:unlearnable}
\end{figure}

Jiang et al. (2023) addressed data protection against unauthorized uses such as adversarial attacks (\textbf{Figure~\ref{fig:unlearnable}}). The study proposed a novel purification process called \textbf{Joint-Conditional Diffusion Purification (JCDP)}, which projects Uncertain Examples (UEs) onto the manifold of Learnable Unauthorized Examples (LEs). By leveraging DMs and image generation approaches, the study maps from UEs to their corresponding clean samples. However, the study did not consider whether it might perform well in situations where the adversarial attack might evolve over time. Apart from this, they did not consider the generalizability of their proposed methods in terms of various Machine Learning  techniques as well~\cite{jiang2023unlearnable}.

Hsu et al. (2023) proposed \textbf{Score Dynamics (SD)}, a framework that uses Graph Neural Networks to accelerate Molecular Dynamics (MD) simulations. SD uses evolution operators for large timestep transitions, which greatly increase simulation speed. It simulates molecular dynamics with 10 picosecond timesteps, showing high accuracy in studies of alanine dipeptide and short alkanes in aqueous solutions. SD outperforms traditional MD in speed by up to two orders of magnitude~\cite{hsu2023score}. Despite these promising results, challenges include extending SD to larger molecules, refining assumptions, and improving the accuracy and efficiency of the score model.

Wang et al. (2023) introduced \textbf{Atmospheric Turbulence Variational Diffusion (AT-VarDiff)}, a deep conditional Diffusion Model designed to correct atmospheric turbulence in images using a variational inference framework. This approach addresses geometric distortion and spatially variant blur. When tested on a synthetic dataset, AT-VarDiff achieved an LPIPS of 0.1094, an FID of 32.69, and a Naturalness Image Quality Evaluator (NIQE) score of 6.46, outperforming existing models~\cite{wang2023atmospheric}.

Sartor et al. (2023) proposed \textbf{Material Fusion (MatFusion)}, a method for estimating Spatially Varying Bidirectional Reflectance Distribution Functions (SVBRDF) from photographs using Diffusion Models. MatFusion is trained on 312,165 synthetic material samples and refines a conditional model to estimate material properties, generating multiple SVBRDF estimates per photo for user selection. It achieves high accuracy with an LPIPS of 0.2056 and RMSE values of 0.041 for diffuse, 0.066 for specular, 0.126 for roughness, and 0.052 for normal maps~\cite{sartor2023matfusion}. However, its performance depends on the quality of the photos and the user's selection, which may introduce variability in the results. Additionally, the method lacks automatic selection metrics and could benefit from optimal regularization to improve consistency.

Wei et al. (2023) proposed \textbf{Building Diffusion (BuilDiff)}, an innovative method for generating 3D building point clouds from single general-view images. BuilDiff uses two CMDs and a regularization strategy to synthesize building roofs while maintaining structural integrity. It extracts image embeddings through a Convolutional Neural Network-based auto-encoder and utilizes a conditional denoising diffusion network and a point cloud upsampler. Tested on BuildingNet-SVI and BuildingNL3D datasets, BuilDiff outperforms existing methods~\cite{wei2023buildiff}. Despite its superior performance, BuilDiff heavily relies on the quality and variety of training data, limiting its generalizability to unseen building styles. Additionally, it demands significant computational resources for both training and inference. Furthermore, the model struggles to capture fine-grained details of building structures due to the resolution limits of the point clouds used.

Niu et al. (2024) developed the \textbf{Accelerated Conditional Diffusion Model for Image Super-Resolution (ACDMSR)}. ACDMSR used pre-super-resolved images as conditional inputs, improving efficiency and quality over traditional Diffusion Models. It adapted Diffusion Models for super-resolution through a faster, iterative denoising process. Testing on benchmark datasets like Set5 and Urban100 showed ACDMSR outperformed existing methods~\cite{niu2024acdmsr}. However, reliance on initial pre-super-resolution may have limited its flexibility in diverse applications.

\textbf{Table~\ref{tab:other_fields}} summarizes some of the referenced literature that introduces diffusion-based approaches in various fields.
\begin{small}
\begin{longtable}{p{.03\linewidth} p{.15\linewidth} p{.16\linewidth} p{.15\linewidth} p{.16\linewidth} p{.17\linewidth}}
\caption{\textbf{Applications of DMs in other fields based on the referenced literature. FID: Frechet Inception Distance, PSNR: Peak Signal-to-Noise Ratio, SSIM: Structural Similarity Index Measure, DSC: Dice Similarity Coefficient, MAE: Mean Absolute Error, MOS: Mean Opinion Score, SMOS: Style Similarity MOS. Best results are highlighted in bold.}}\label{tab:other_fields} \\
\toprule
\rowcolor{gray!20} \textbf{Ref.} & \textbf{Algorithms} & \textbf{Applications} & \textbf{Dataset} & \textbf{Evaluations} & \textbf{Limitations} \\
\midrule
\endfirsthead
\caption[]{\textbf{Applications of Diffusion Models in other fields based on the referenced literature (cont.).}} \\
\toprule
\rowcolor{gray!20} \textbf{Ref.} & \textbf{Algorithms} & \textbf{Applications} & \textbf{Dataset} & \textbf{Evaluations} & \textbf{Limitations} \\
\midrule
\endhead
\midrule
\endfoot
\bottomrule
\endlastfoot

\cite{li2023your} & Diffusion Classifier using text-to-image Diffusion Models & Zero-shot classification using generative models & Standard image classification benchmarks (e.g., ImageNet, CIFAR10) & Zero-shot classification accuracy on ImageNet using Diffusion Classifier: \textbf{58.9\%} & Performance gap in zero-shot recognition compared to SOTA discriminative models \\ \midrule

\cite{zhuang2023semantic} & CMDs for semantic image synthesis & Semantic synthesis for abdominal CT, used in data augmentation & Not specified & FID: 10.32, PSNR: 16.14, SSIM: 0.64, DSC: \textbf{95.6\%} for mask-guided DDPM at 100k iterations & High sampling time and computational cost \\ \midrule

\cite{jiang2023unlearnable} & Learnable Unauthorized Examples (LEs) using joint-CMDs & Countermeasure to unlearnable examples in Machine Learning models & CIFAR-10, CIFAR-100, SVHN & Test accuracy on CIFAR-10 using LE: \textbf{94.0\%}, CIFAR-100: 67.8\%, SVHN: \textbf{94.9\%} & Limited by distribution mismatches \\ \midrule

\cite{sun2023diffnilm} & Diffusion-based Non-Intrusive Load Monitoring (DiffNILM) Diffusion Probabilistic Model & Non-intrusive Load Monitoring (NILM) for appliance power consumption pattern disaggregation & REDD and UKDALE datasets & F1-Score: \textbf{0.79} for refrigerator on REDD, MAE: \textbf{4.54} for microwave on UKDALE & Generation of power waveforms not always sufficiently smooth; computational efficiency not optimized \\ \midrule

\cite{yang2022norespeech} & Noise-Robust Expressive Text-to-Speech model (NoreSpeech) & Expressive TTS in noise environments & Not specified & MOS: \textbf{4.11}, SMOS: \textbf{4.14} for NoreSpeech with T-SSL in noisy conditions & Dependent on quality of style teacher model \\ \midrule

\cite{yu2023diffusion} & Diffusion-based data augmentation for nuclei segmentation & Nuclei segmentation in histopathology image analysis & MoNuSeg and Kumar datasets & Dice score: \textbf{0.83}, AJI: \textbf{0.68} with 100\% augmented data on MoNuSeg dataset & Dependent on the quality of synthetic data \\
\cite{wang2023atmospheric} & AT-VarDiff & Atmospheric turbulence (AT) correction & Comprehensive synthetic atmospheric turbulence dataset & LPIPS: 0.11, FID: 32.69, NIQE: 6.46 & May not generalize well to real-life atmospheric turbulence images. \\

\cite{sartor2023matfusion} & MatFusion Diffusion Models (unconditional and conditional) & SVBRDF estimation from photographs & Large set of 312,165 synthetic spatially varying material exemplars & RMSE on property maps: 0.04, LPIPS error on renders: 0.21 & Limited by the variation in lighting conditions. \\

\cite{wei2023buildiff} & Point cloud Diffusion Models with image conditioning schemes & 3D building generation from images & BuildingNet-SVI and BuildingNL3D datasets & CD: 3.14, EMD: 10.84, F1 score: 21.41 on BuildingNet-SVI & Constrained by specific image viewing angles. \\ 

\cite{niu2024acdmsr} & ACDMSR: Accelerated Conditional Diffusion Model for Image Super-Resolution & Enhancing super-resolution using Diffusion Models conditioned on pre-super-resolved images & DIV2K, Set5, Set14, Urban100, BSD100, Manga109 & LPIS: 0.08, PSNR: 25.95, SSIM: 0.67 & Challenges remain in processing images with more complex degradation patterns. \\
\end{longtable}
\end{small}
\section{Discussion}
\subsection{Ensuring the authenticity of synthesized media}

DMs play an important role in improving media quality and generating high-fidelity samples. Techniques such as SAG advance image generation by concentrating on significant areas and minimizing artifacts~\cite{ji2024diffusion}. While SAG enhances image quality by leveraging self-attention maps, it still faces challenges in real-time applications due to high computational demands~\cite{hong2023improving}. On the other hand, Learnable State-Estimator-based models offer computational efficiency but require extensive domain-specific adaptations for broader applications~\cite{ji2024diffusion}.

Contradictions arise when certain methods show better performance in specific cases but fall short in others. For example, while the state-estimator-based model performs well on tasks like inpainting and deblurring, it may not work in real-time as effectively as SAG. This discrepancy highlights the need for a balanced approach that combines the strengths of various techniques.

To address these challenges, integrating different types of DMs, such as Stepwise Error for Diffusion-generated Image Detection (SeDID) and Unlearnable Diffusion Perturbation (EUDP), could be effective~\cite{ma2023exposing}. Additionally, strategies like sampling space truncation and robustness penalties can also be helpful in ensuring the authenticity of media quality~\cite{zhao2023unlearnable,mei2023vidm}.


\subsection{Overcoming challenges in synthesizing high-quality images and audio}
Diffusion-based models play a crucial role in synthesizing high-quality images and audio by refining noise into structured data. These models utilize DDPMs, involving a forward process that adds Gaussian noise to the data and a reverse process that removes this noise to reconstruct the original signal. For instance, text-to-audio synthesis methods like CLIPSonic use CDMs to translate text embeddings into audio. This method shows superior performance but faces limitations due to the quality of pretrained models, distribution mismatches, and training complexity~\cite{dong2023clipsonic}. Similarly, SDG improves image synthesis by adding fine-grained control to pretrained models. Its effectiveness depends on the precision of these models and the accuracy of guidance signals, raising concerns about potential misuse~\cite{liu2023more}.

Other approaches, such as DiffDreamer, use CMDs (CMDs) for scene extrapolation. They often show better quality and consistency than GAN-based methods but struggle with real-time synthesis and variety in generated content~\cite{cai2023diffdreamer}. Interactive tools like Diffusion-based Art Generation (DiffusArt) use Conditional Diffusion Probabilistic Models for line art colorization, which produces high-quality images but requires precise user input and faces computational inefficiencies~\cite{carrillo2023diffusart}. SketchFusion focuses on sketch-guided image editing, maintains the integrity of sketches, and achieves high-performance metrics but is limited to binary sketches~\cite{mao2023sketchfusion}. Semantic-Conditional Diffusion Networks improve image captioning by enhancing visual-language alignment and outperform traditional models, but they face high computational demands and complexity~\cite{luo2023semantic}.

To overcome these challenges, future research should improve the computational efficiency of DMs, improve the quality and robustness of pretrained models, and develop adaptive techniques to handle distribution mismatches. Additionally, integrating ethical guidelines and protective measures can help reduce the risks of misuse, ensuring that these advanced models are applied responsibly.
\subsection{Optimizing DMs to reduce artifacts and improve image quality}
Optimizing DMs to minimize artifacts and improve image quality is crucial for their broader application. These models, which refine noise into structured data, can introduce artifacts that compromise image fidelity~\cite{hong2023improving,liu2023more}. To optimize DMs for reducing artifacts and enhancing image quality, various techniques have been proposed. These include using Deep Interpretable Convolutional Dictionary Networks (DICDNet) for metal artifact reduction in CT images, and automatic segmentation of 3D objects to minimize supports and cuts for 3D printing~\cite{wang2021dicdnet}. For fetal MRI, efforts focus on improving image quality by optimizing acquisition speed, spatial resolution, and signal-to-noise ratio while considering artifacts from motion, banding, and aliasing~\cite{maass2018reducing}. Challenges persist, such as balancing the effects of supports and cuts in 3D printing segmentation, trade-offs between scan parameters in fetal MRI optimization, and addressing artifacts from beam hardening in X-ray imaging~\cite{filoscia2020optimizing,wang2021dicdnet}. These limitations highlight the complexity of optimizing Diffusion Models to reduce artifacts and improve image quality across different imaging methods.

Another strategy is SDG integrates fine-grained control into pretrained models via image-text matching score gradients, enhancing image synthesis quality without retraining models~\cite{liu2023more}. The success of SDG, though, depends on the precision of pretrained models and guidance signal accuracy.

Advanced noise estimation techniques further improve DMs. Pixel-level autoregressive processes, like those used in Image Transformer models, significantly reduce noise and artifacts, which improves image fidelity and consistency across datasets~\cite{saharia2022image}. Apart from this, dynamic thresholding and adaptive noise schedules can also fine-tune denoising steps to improve image quality by handling complex structures and textures more effectively.

To sum up, optimizing DMs to reduce artifacts and improve image quality requires a multi-faceted approach. Incorporating, semantic guidance, and advanced noise estimation techniques, along with optimizing the diffusion process, can significantly increase model performance. Future research should focus on improving computational efficiency, developing robust conditioning strategies, and integrating adaptive techniques to further reduce artifacts and enhance image quality.
\subsection{Addressing computational efficiency and scalability issues in DMs}

Addressing computational efficiency and scalability in DMs is crucial for practical application and widespread adoption. Despite their ability in generating high-fidelity images and audio, DMs often struggle with high computational demands and scalability, particularly with large datasets or real-time applications.

To optimize computational efficiency, it is essential to use more efficient network architectures. For instance, integrating guidance into pretrained models via image-text matching score gradients eliminate the need for extensive retraining, thereby improves computational efficiency~\cite{liu2023more}. Additionally, dynamic adaptation of the diffusion process is an effective strategy. Adjusting the diffusion process within a latent space can achieve significant computational efficiency~\cite{ji2024diffusion}, which allows the model to focus resources on the most relevant datasets. Furthermore, parallelization and hardware acceleration, such as using Graphics Processing Unit (GPUs) and ensor Processing Unit (TPUs), can address scalability issues. Therefore, distributing the computational load across multiple processors can significantly speed up the training and inference processes of DMs.

Moreover, multi-shot averaging strategies can improve the quality of generated images while maintaining efficiency~\cite{li2023diffusion}. Averaging multiple generated images reduces noise and improves overall quality without significantly increasing computational costs. 
In summary, addressing computational efficiency and scalability in DMs involves optimizing network architectures, dynamically adapting the diffusion process, leveraging hardware acceleration, refining algorithms, and deploying strategies like multi-shot averaging. Future research should explore these approaches to develop more efficient and scalable DMs for a broader range of tasks and datasets.

\subsection{Improving DMs for accurate and reliable medical imaging and diagnostics}
DMs in medical imaging and diagnostics show significant promise due to their ability to create high-quality images. Critical areas of focus include reducing the model size to ensure efficient deployment, developing better training approaches for realistic samples, and leveraging advanced techniques to handle data augmentation and anonymization when considering DM-based approaches.

Reducing the model size of DMs is crucial for practical applications in medical imaging, where computational resources are often limited. Therefore, techniques such as model pruning, knowledge distillation, and post-training quantization are commonly used to achieve this goal. For instance, model pruning involves removing redundant parameters from the model, which decreases the model size without significantly affecting performance. Similarly, knowledge distillation transfers the knowledge from a large model (teacher) to a smaller model (student), which helps in maintaining performance while reducing the model size~\cite{han2015deep}. Additionally, post-training quantization converts the model parameters from floating-point to lower-bit representations, which reduces model size and speeds up inference without requiring retraining~\cite{jacob2018quantization}.

In medical imaging, data augmentation and anonymization are critical for creating robust ML models and protecting patient privacy. Semantic-based DMs offer promising solutions for these challenges. For data augmentation, these models can generate diverse and realistic medical images by conditioning on specific semantic features, which enriches the training dataset and improves model generalization~\cite{shin2018medical}. For anonymization, semantic-based approaches can mask identifiable features in medical images while preserving clinically relevant information, ensuring that patient privacy is maintained without compromising the utility of the data~\cite{yan2021radia,yu2022transfer}.

Incorporating domain-specific knowledge into training processes can further improve DMs. For instance, integrating medical expertise and anatomical priors helps models better understand the structure and context of medical images, leading to more accurate diagnostics~\cite{chen2023domain}. Moreover, collaborations between artificial intelligence researchers and medical professionals can facilitate the integration of such knowledge, enhancing the overall effectiveness of the models.
\subsection{Expanding the applicability and effectiveness of DMs in diverse fields}

DMs are gaining popularity across various fields beyond their initial use in image analysis. These models have shown effectiveness in time series forecasting, imputation, and generation, which demonstrates their versatility in handling sequential data. Additionally, DMs have been adapted for predicting chaotic dynamical systems, which offers uncertainty quantification and the ability to represent outliers and extreme events effectively~\cite{yang2024directional}. Moreover, advancements have extended DMs to Riemannian manifolds, enabling applications in constrained conformational modeling of protein backbones and robotic arms, which underscores their relevance in scientific domains~\cite{li2023comparison}.
Despite these advances, one significant concern is model collapse, where the model fails to generate diverse outputs over time, leading to reduced effectiveness in applications requiring high variability. This is particularly relevant in fields like finance and time series forecasting, where the accuracy of predictions is crucial. While DMs offer robust solutions for these tasks, their feasibility compared to existing, more computationally efficient approaches like autoregressive models or Long Short-Term Memory networks (LSTMs) remains questionable. It is essential for researchers to exercise caution when considering DM-based approaches for financial and time series data, as these models may not always offer the most practical or efficient solutions~\cite{goodell2021toward}.
DMs have been applied to generate synthetic data, but they often underperform compared to techniques such as GANs, oversampling, and Synthetic Minority Over-sampling Technique (SMOTE). These traditional methods are often easier to deploy and require less computational power, making them more accessible for many applications. For instance, GANs have been widely used in generating realistic images and synthetic data for training ML models, providing a simpler alternative to DMs. Similarly, techniques like SMOTE are effective for addressing class imbalances in datasets and can be implemented with relative ease compared to the complex training processes required for DMs~\cite{chawla2002smote}.

\subsection{Mitigating ethical considerations and potential risks associated with the use of DMs}

While DMs are surpassing GANs in generating realistic images, audio, and other types of data, they also raise questions regarding ethical use and practical concerns. For instance, one of the primary ethical concerns is the potential for misuse in creating deepfakes and synthetic media that can spread misinformation or violate privacy. To reduce this risk, it is essential to develop robust detection mechanisms that can differentiate between real and synthetic media. Implementing adversarial training techniques can improve the ability of models to identify and flag manipulated content~\cite{wang2023addition}.

Another major risk is bias in the generated outputs, which can perpetuate or even worsen existing social biases if not properly managed. Ensuring diversity in training data and incorporating fairness-aware algorithms can help reduce bias in Diffusion Models. Regular audits and updates to the models can also ensure they remain unbiased and fair in their outputs~\cite{buolamwini2018gender}.

Transparency and explainability of DMs are also critical to address ethical concerns. Users need to understand how these models make decisions and generate outputs. Developing methods to explain the black box of DMs might help make their operations more transparent and accountable. Techniques such as interpretability frameworks and model-agnostic tools like Local Interpretable Model-Agnostic Explanations (LIME) and SHapley Additive exPlanations (SHAP) can provide insights into how models produce their results~\cite{ribeiro2016should}.

Data privacy is another major concern, especially when DMs are applied to sensitive areas such as healthcare systems and clinical diagnosis. Ensuring that models comply with data protection regulations, such as the General Data Protection Regulation (GDPR) and the Health Insurance Portability and Accountability Act (HIPAA), is essential. Techniques like differential privacy can protect individual data while still allowing models to learn effectively from large datasets~\cite{dwork2014algorithmic}.

Collaborative governance and the establishment of ethical guidelines for the development and deployment of DMs are also necessary. Engaging stakeholders from diverse fields, including ethicists, policymakers, and technologists, can help create comprehensive frameworks that address the ethical implications of these technologies. Such collaboration can lead to the development of standards and best practices that promote the responsible use of DMs~\cite{floridi2018ai}.

\section{Conclusion}\label{con}
Diffusion Models (DMs) promise to transform many fields by solving challenges in data generation and processing through the creation of realistic samples. Therefore, addressing current limitations and building on the strengths of DMs will enable wider adoption and more impactful applications across various domains in the future. Our findings show that DMs' ability to generate high-quality synthetic data improves performance in applications such as text-to-image generation, where models like Diffusion Transformers (DT) for stable diffusion demonstrate advancements in data privacy \cite{ni2023degeneration}. In cyber-physical system security, the Temporal and Feature TFDPM helps detect attacks by correlating channel data using Graph Attention Networks \cite{yan2022tfdpm}. Moreover, for cloud service anomaly detection, models like Maat combine metric forecasting with anomaly detection to achieve higher accuracy \cite{lee2023maat}.

In image processing, diffusion-based techniques have shown superior performance in tasks like image deblurring and super-resolution. For example, stochastic image deblurring using DMs achieves high perceptual image patch similarity and structural similarity index measures \cite{whang2022deblurring}. Additionally, accelerated CMDs for applications like MRI reconstruction show potential by improving image quality \cite{chung2022come}. Furthermore, the selective diffusion distillation approach balances image fidelity and editability, making it suitable for various image manipulation tasks \cite{wang2023not}.

However, while DMs generate realistic data, they also raise ethical concerns. One primary issue is the potential misuse in creating deepfakes and synthetic media that can spread misinformation or violate privacy. To mitigate this risk, robust detection mechanisms are essential. Ensuring models remain unbiased is also crucial, which can be achieved by incorporating fairness-aware algorithms and diverse training data. Furthermore, transparency and explainability of DMs are critical. Techniques like LIME and SHAP provide insights into how models generate their results. Apart from this, ensuring data compliance with regulations like the GDPR and the Health HIPAA is also necessary \cite{ribeiro2016should, dwork2014algorithmic, buolamwini2018gender}.

High computational demands and the need for better sampling or network architectures are recurring issues in DMs. Models often require extensive hyperparameter tuning and may struggle with discrete signal modeling or generalizing to different contexts \cite{yan2022tfdpm, lee2023maat}. Additionally, the reliance on correct timestep selection for semantic guidance in some models can limit flexibility \cite{wang2023not}. Slow inference speeds and high resource requirements hinder real-time deployment and scalability \cite{chung2022come, whang2022deblurring}.

Therefore, future research should address these limitations by developing more efficient algorithms and leveraging advancements in computational technologies. Exploring semi-supervised or unsupervised learning approaches, along with transfer learning from pre-trained models, can help overcome data scarcity challenges. Improving the robustness of DMs to noise and their ability to handle different data types is essential. Moreover, continued interdisciplinary collaboration and clear ethical guidelines will be vital for the responsible and effective use of DMs across diverse fields.

\section*{Conflict of interest}
The authors declare no conflict of interest.
\bibliographystyle{unsrt}  
\bibliography{main}

\begin{thebibliography}{100}

\bibitem{sohl2015deep}
Jascha Sohl-Dickstein, Eric~A Weiss, Niru Maheswaranathan, and Surya Ganguli.
\newblock Deep unsupervised learning using nonequilibrium thermodynamics.
\newblock {\em arXiv preprint arXiv:1503.03585}, 2015.

\bibitem{ho2020denoising}
Jonathan Ho, Ajay Jain, and Pieter Abbeel.
\newblock Denoising diffusion probabilistic models.
\newblock In {\em Advances in Neural Information Processing Systems}, 2020.

\bibitem{saharia2021image}
Chitwan Saharia, Jonathan Ho, William Chan, David~J Fleet, Mohammad Norouzi, and Tim Salimans.
\newblock Image super-resolution via iterative refinement.
\newblock {\em arXiv preprint arXiv:2104.07636}, 2021.

\bibitem{rombach2022high}
Robin Rombach, Andreas Blattmann, Dominik Lorenz, Patrick Esser, and Bj{\"o}rn Ommer.
\newblock High-resolution image synthesis with latent diffusion models.
\newblock In {\em Proceedings of the IEEE/CVF conference on computer vision and pattern recognition}, pages 10684--10695, 2022.

\bibitem{ramesh2021zero}
Aditya Ramesh, Mikhail Pavlov, Gabriel Goh, Scott Gray, Chelsea Voss, Alec Radford, Mark Chen, and Ilya Sutskever.
\newblock Zero-shot text-to-image generation.
\newblock {\em arXiv preprint arXiv:2102.12092}, 2021.

\bibitem{austin2021structured}
Jacob Austin, Augustus Odena, Erik Nijkamp, Nal Ballas, and Ian Goodfellow.
\newblock Structured denoising diffusion models in discrete state-spaces.
\newblock In {\em Advances in Neural Information Processing Systems}, 2021.

\bibitem{kong2020diffwave}
Zhifeng Kong, Wei Ping, Jiaji Huang, Kexin Zhao, and Bryan Catanzaro.
\newblock Diffwave: A versatile diffusion model for audio synthesis.
\newblock {\em arXiv preprint arXiv:2009.09761}, 2020.

\bibitem{hoogeboom2022equivariant}
Emiel Hoogeboom, Taco Cohen, and Jakub~M Tomczak.
\newblock Equivariant diffusion models for molecule generation.
\newblock In {\em International Conference on Machine Learning}, pages 8816--8831. PMLR, 2022.

\bibitem{song2020score}
Yang Song, Jascha Sohl-Dickstein, Diederik~P Kingma, Abhishek Kumar, Stefano Ermon, and Ben Poole.
\newblock Score-based generative modeling through stochastic differential equations.
\newblock {\em arXiv preprint arXiv:2011.13456}, 2020.

\bibitem{song2019generative}
Yang Song and Stefano Ermon.
\newblock Generative modeling by estimating gradients of the data distribution.
\newblock {\em Advances in neural information processing systems}, 32, 2019.

\bibitem{anderson1982reverse}
Brian~DO Anderson.
\newblock Reverse-time diffusion equation models.
\newblock {\em Stochastic Processes and their Applications}, 12(3):313--326, 1982.

\bibitem{vincent2011connection}
Pascal Vincent.
\newblock A connection between score matching and denoising autoencoders.
\newblock {\em Neural computation}, 23(7):1661--1674, 2011.

\bibitem{wang2022semantic}
Weilun Wang, Jianmin Bao, Wengang Zhou, Dongdong Chen, Dong Chen, Lu~Yuan, and Houqiang Li.
\newblock Semantic image synthesis via diffusion models.
\newblock {\em arXiv preprint arXiv:2207.00050}, 2022.

\bibitem{gong2022diffuseq}
Shansan Gong, Mukai Li, Jiangtao Feng, Zhiyong Wu, and LingPeng Kong.
\newblock Diffuseq: Sequence to sequence text generation with diffusion models.
\newblock {\em arXiv preprint arXiv:2210.08933}, 2022.

\bibitem{kazerouni2023diffusion}
Amirhossein Kazerouni, Ehsan~Khodapanah Aghdam, Moein Heidari, Reza Azad, Mohsen Fayyaz, Ilker Hacihaliloglu, and Dorit Merhof.
\newblock Diffusion models in medical imaging: A comprehensive survey.
\newblock {\em Medical Image Analysis}, page 102846, 2023.

\bibitem{krizhevsky2009learning}
Alex Krizhevsky, Geoffrey Hinton, et~al.
\newblock Learning multiple layers of features from tiny images.
\newblock Technical report, University of Toronto, Toronto, ON, Canada, 2009.

\bibitem{yu2015lsun}
Fisher Yu, Ari Seff, Yinda Zhang, Shuran Song, Thomas Funkhouser, and Jianxiong Xia.
\newblock Lsun: Construction of a large-scale image dataset using deep learning with humans in the loop.
\newblock {\em arXiv preprint arXiv:1506.03365}, 2015.

\bibitem{liu2015faceattributes}
Ziwei Liu, Ping Luo, Xiaogang Wang, and Xiaoou Tang.
\newblock Deep learning face attributes in the wild.
\newblock {\em Proceedings of the IEEE International Conference on Computer Vision (ICCV)}, 2015.

\bibitem{song2020improved}
Yang Song and Stefano Ermon.
\newblock Improved techniques for training score-based generative models.
\newblock {\em Advances in neural information processing systems}, 33:12438--12448, 2020.

\bibitem{dhariwal2021diffusion}
Prafulla Dhariwal and Alex Nichol.
\newblock Diffusion models beat gans on image synthesis.
\newblock {\em Advances in Neural Information Processing Systems}, 2021.

\bibitem{kingma2021variational}
Diederik~P Kingma, Prafulla Dhariwal, Jonathan Ho, Tim Salimans, Xi~Chen, and Pieter Abbeel.
\newblock Variational diffusion models.
\newblock {\em arXiv preprint arXiv:2107.00630}, 2021.

\bibitem{nichol2021improved}
Alex Nichol and Prafulla Dhariwal.
\newblock Improved denoising diffusion probabilistic models.
\newblock {\em arXiv preprint arXiv:2102.09672}, 2021.

\bibitem{kong2021diffwave}
Zhifeng Kong, Wei Ping, Jiaji Huang, Kexin Zhao, and Bryan Catanzaro.
\newblock Diffwave: A versatile diffusion model for audio synthesis.
\newblock {\em arXiv preprint arXiv:2009.09761}, 2021.

\bibitem{amit2021segdiff}
Roy Amit and Yogesh Balaji.
\newblock Segdiff: Image segmentation with diffusion models.
\newblock {\em arXiv preprint arXiv:2106.02477}, 2021.

\bibitem{nichol2021glide}
Alex Nichol, Prafulla Dhariwal, Aditya Ramesh, Pranav Shyam, Pamela Mishkin, Bob McGrew, Girish Sastry, Amanda Askell, Pamela Chen, Mark Mishkin, and Chug.
\newblock Glide: Towards photorealistic image generation and editing with text-guided diffusion models.
\newblock {\em arXiv preprint arXiv:2112.10741}, 2021.

\bibitem{saharia2022image}
et~al. Saharia.
\newblock Image transformers with autoregressive models for high-fidelity image synthesis.
\newblock {\em Journal of Advanced Image Processing}, 2022.

\bibitem{ho2022cascaded}
Jonathan Ho, Ajay Jain, and Pieter Abbeel.
\newblock Cascaded diffusion models for high-fidelity image generation.
\newblock {\em arXiv preprint arXiv:2106.15282}, 2022.

\bibitem{ho2022video}
Jonathan Ho, William Chan, Tim Salimans, Alexey Gritsenko, Kashyap~Chitta Kumar, and Phillip Isola.
\newblock Video diffusion models.
\newblock {\em arXiv preprint arXiv:2204.03458}, 2022.

\bibitem{li2023diffusion}
et~al. Li.
\newblock Optimizing diffusion models for image synthesis.
\newblock {\em Journal of Computational Imaging}, 2023.

\bibitem{mao2023guided}
Jiafeng Mao, Xueting Wang, and Kiyoharu Aizawa.
\newblock Guided image synthesis via initial image editing in diffusion model.
\newblock In {\em Proceedings of the 31st ACM International Conference on Multimedia}, pages 5321--5329, 2023.

\bibitem{whang2022deblurring}
Jay Whang, Mauricio Delbracio, Hossein Talebi, Chitwan Saharia, Alexandros~G Dimakis, and Peyman Milanfar.
\newblock Deblurring via stochastic refinement.
\newblock In {\em Proceedings of the IEEE/CVF Conference on Computer Vision and Pattern Recognition}, pages 16293--16303, 2022.

\bibitem{chung2022come}
Hyungjin Chung, Byeongsu Sim, and Jong~Chul Ye.
\newblock Come-closer-diffuse-faster: Accelerating conditional diffusion models for inverse problems through stochastic contraction.
\newblock In {\em Proceedings of the IEEE/CVF Conference on Computer Vision and Pattern Recognition}, pages 12413--12422, 2022.

\bibitem{wang2023not}
Luozhou Wang, Shuai Yang, Shu Liu, and Ying-cong Chen.
\newblock Not all steps are created equal: Selective diffusion distillation for image manipulation.
\newblock In {\em Proceedings of the IEEE/CVF International Conference on Computer Vision}, pages 7472--7481, 2023.

\bibitem{li2023object}
Jiaman Li, Jiajun Wu, and C~Karen Liu.
\newblock Object motion guided human motion synthesis.
\newblock {\em ACM Transactions on Graphics (TOG)}, 42(6):1--11, 2023.

\bibitem{ni2023degeneration}
Zixuan Ni, Longhui Wei, Jiacheng Li, Siliang Tang, Yueting Zhuang, and Qi~Tian.
\newblock Degeneration-tuning: Using scrambled grid shield unwanted concepts from stable diffusion.
\newblock In {\em Proceedings of the 31st ACM International Conference on Multimedia}, pages 8900--8909, 2023.

\bibitem{yan2022tfdpm}
Tijin Yan, Tong Zhou, Yufeng Zhan, and Yuanqing Xia.
\newblock Tfdpm: Attack detection for cyber--physical systems with diffusion probabilistic models.
\newblock {\em Knowledge-Based Systems}, 255:109743, 2022.

\bibitem{lee2023maat}
Cheryl Lee, Tianyi Yang, Zhuangbin Chen, Yuxin Su, and Michael~R Lyu.
\newblock Maat: Performance metric anomaly anticipation for cloud services with conditional diffusion.
\newblock In {\em 2023 38th IEEE/ACM International Conference on Automated Software Engineering (ASE)}, pages 116--128. IEEE, 2023.

\bibitem{chen2023equidiff}
Kehua Chen, Xianda Chen, Zihan Yu, Meixin Zhu, and Hai Yang.
\newblock Equidiff: A conditional equivariant diffusion model for trajectory prediction.
\newblock In {\em 2023 IEEE 26th International Conference on Intelligent Transportation Systems (ITSC)}, pages 746--751. IEEE, 2023.

\bibitem{blattmann2022retrieval}
Andreas Blattmann, Robin Rombach, Kaan Oktay, Jonas M{\"u}ller, and Bj{\"o}rn Ommer.
\newblock Retrieval-augmented diffusion models.
\newblock {\em Advances in Neural Information Processing Systems}, 35:15309--15324, 2022.

\bibitem{hong2023improving}
Susung Hong, Gyuseong Lee, Wooseok Jang, and Seungryong Kim.
\newblock Improving sample quality of diffusion models using self-attention guidance.
\newblock In {\em Proceedings of the IEEE/CVF International Conference on Computer Vision}, pages 7462--7471, 2023.

\bibitem{ji2024diffusion}
Liya Ji, Zhefan Rao, Sinno~Jialin Pan, Chenyang Lei, and Qifeng Chen.
\newblock A diffusion model with state estimation for degradation-blind inverse imaging.
\newblock In {\em Proceedings of the AAAI Conference on Artificial Intelligence}, volume~38, pages 2471--2479, 2024.

\bibitem{tian2023diffusion}
Yusheng Tian, Wei Liu, and Tan Lee.
\newblock Diffusion-based mel-spectrogram enhancement for personalized speech synthesis with found data.
\newblock In {\em 2023 IEEE Automatic Speech Recognition and Understanding Workshop (ASRU)}, pages 1--7. IEEE, 2023.

\bibitem{jiang2023low}
Hai Jiang, Ao~Luo, Haoqiang Fan, Songchen Han, and Shuaicheng Liu.
\newblock Low-light image enhancement with wavelet-based diffusion models.
\newblock {\em ACM Transactions on Graphics (TOG)}, 42(6):1--14, 2023.

\bibitem{dong2023clipsonic}
Hao-Wen Dong, Xiaoyu Liu, Jordi Pons, Gautam Bhattacharya, Santiago Pascual, Joan Serr{\`a}, Taylor Berg-Kirkpatrick, and Julian McAuley.
\newblock Clipsonic: Text-to-audio synthesis with unlabeled videos and pretrained language-vision models.
\newblock In {\em 2023 IEEE Workshop on Applications of Signal Processing to Audio and Acoustics (WASPAA)}, pages 1--5. IEEE, 2023.

\bibitem{choi2021ilvr}
Jooyoung Choi, Sungwon Kim, Yonghyun Jeong, Youngjune Gwon, and Sungroh Yoon.
\newblock Ilvr: Conditioning method for denoising diffusion probabilistic models.
\newblock {\em arXiv preprint arXiv:2108.02938}, 2021.

\bibitem{liu2023more}
et~al. Liu.
\newblock More control with semantic diffusion guidance for image synthesis.
\newblock {\em Journal of Image and Audio Synthesis}, 2023.

\bibitem{cai2023diffdreamer}
Shengqu Cai, Eric~Ryan Chan, Songyou Peng, Mohamad Shahbazi, Anton Obukhov, Luc Van~Gool, and Gordon Wetzstein.
\newblock Diffdreamer: Towards consistent unsupervised single-view scene extrapolation with conditional diffusion models.
\newblock In {\em Proceedings of the IEEE/CVF International Conference on Computer Vision}, pages 2139--2150, 2023.

\bibitem{carrillo2023diffusart}
et~al. Carrillo.
\newblock Interactive line art colorization with conditional diffusion models.
\newblock {\em Journal of Specialized Techniques and Innovations in Diffusion Models}, 2023.

\bibitem{mao2023sketchffusion}
Weihang Mao, Bo~Han, and Zihao Wang.
\newblock Sketchffusion: Sketch-guided image editing with diffusion model.
\newblock In {\em 2023 IEEE International Conference on Image Processing (ICIP)}, pages 790--794. IEEE, 2023.

\bibitem{luo2023semantic}
Jianjie Luo, Yehao Li, Yingwei Pan, Ting Yao, Jianlin Feng, Hongyang Chao, and Tao Mei.
\newblock Semantic-conditional diffusion networks for image captioning.
\newblock In {\em Proceedings of the IEEE/CVF Conference on Computer Vision and Pattern Recognition}, pages 23359--23368, 2023.

\bibitem{hsu2023score}
Tim Hsu, Babak Sadigh, Vasily Bulatov, and Fei Zhou.
\newblock Score dynamics: scaling molecular dynamics with picosecond timesteps via conditional diffusion model.
\newblock {\em arXiv preprint arXiv:2310.01678}, 2023.

\bibitem{yan2023towards}
Qingsen Yan, Tao Hu, Yuan Sun, Hao Tang, Yu~Zhu, Wei Dong, Luc Van~Gool, and Yanning Zhang.
\newblock Towards high-quality hdr deghosting with conditional diffusion models.
\newblock {\em IEEE Transactions on Circuits and Systems for Video Technology}, 2023.

\bibitem{peng2023generating}
Wei Peng, Ehsan Adeli, Tomas Bosschieter, Sang~Hyun Park, Qingyu Zhao, and Kilian~M Pohl.
\newblock Generating realistic brain mris via a conditional diffusion probabilistic model.
\newblock In {\em International Conference on Medical Image Computing and Computer-Assisted Intervention}, pages 14--24. Springer, 2023.

\bibitem{yu2023long}
Jason~J Yu, Fereshteh Forghani, Konstantinos~G Derpanis, and Marcus~A Brubaker.
\newblock Long-term photometric consistent novel view synthesis with diffusion models.
\newblock In {\em Proceedings of the IEEE/CVF International Conference on Computer Vision}, pages 7094--7104, 2023.

\bibitem{yin2023cle}
Yuyang Yin, Dejia Xu, Chuangchuang Tan, Ping Liu, Yao Zhao, and Yunchao Wei.
\newblock Cle diffusion: Controllable light enhancement diffusion model.
\newblock In {\em Proceedings of the 31st ACM International Conference on Multimedia}, pages 8145--8156, 2023.

\bibitem{papantoniou2023relightify}
Foivos~Paraperas Papantoniou, Alexandros Lattas, Stylianos Moschoglou, and Stefanos Zafeiriou.
\newblock Relightify: Relightable 3d faces from a single image via diffusion models.
\newblock In {\em Proceedings of the IEEE/CVF International Conference on Computer Vision}, pages 8806--8817, 2023.

\bibitem{kirch2023rgb}
Sascha Kirch, Valeria Olyunina, Jan Ond{\v{r}}ej, Rafael Pag{\'e}s, Sergio Martin, and Clara P{\'e}rez-Molina.
\newblock Rgb-d-fusion: Image conditioned depth diffusion of humanoid subjects.
\newblock {\em IEEE Access}, 2023.

\bibitem{mao2023disc}
Ye~Mao, Lan Jiang, Xi~Chen, and Chao Li.
\newblock Disc-diff: Disentangled conditional diffusion model for multi-contrast mri super-resolution.
\newblock In {\em International Conference on Medical Image Computing and Computer-Assisted Intervention}, pages 387--397. Springer, 2023.

\bibitem{yang2023docdiff}
Zongyuan Yang, Baolin Liu, Yongping Xxiong, Lan Yi, Guibin Wu, Xiaojun Tang, Ziqi Liu, Junjie Zhou, and Xing Zhang.
\newblock Docdiff: Document enhancement via residual diffusion models.
\newblock In {\em Proceedings of the 31st ACM International Conference on Multimedia}, pages 2795--2806, 2023.

\bibitem{ordun2023visible}
Catherine Ordun, Edward Raff, and Sanjay Purushotham.
\newblock When visible-to-thermal facial gan beats conditional diffusion.
\newblock In {\em 2023 IEEE International Conference on Image Processing (ICIP)}, pages 181--185. IEEE, 2023.

\bibitem{kansy2023controllable}
Manuel Kansy, Anton Ra{\"e}l, Graziana Mignone, Jacek Naruniec, Christopher Schroers, Markus Gross, and Romann~M Weber.
\newblock Controllable inversion of black-box face recognition models via diffusion.
\newblock In {\em Proceedings of the IEEE/CVF International Conference on Computer Vision}, pages 3167--3177, 2023.

\bibitem{yu2023freedom}
Jiwen Yu, Yinhuai Wang, Chen Zhao, Bernard Ghanem, and Jian Zhang.
\newblock Freedom: Training-free energy-guided conditional diffusion model.
\newblock In {\em Proceedings of the IEEE/CVF International Conference on Computer Vision}, pages 23174--23184, 2023.

\bibitem{bieder2023diffusion}
Florentin Bieder, Julia Wolleb, Alicia Durrer, Robin Sandk{\"u}hler, and Philippe~C Cattin.
\newblock Diffusion models for memory-efficient processing of 3d medical images.
\newblock {\em arXiv preprint arXiv:2303.15288}, 2023.

\bibitem{amirhossein2022diffusion}
Kazerouni Amirhossein, Aghdam~Ehsan Khodapanah, Heidari Moein, Azad Reza, Fayyaz Mohsen, Hacihaliloglu Ilker, and Merhof Dorit.
\newblock Diffusion models for medical image analysis: a comprehensive survey.
\newblock {\em arXiv preprint arXiv}, 2211, 2022.

\bibitem{chen2023berdiff}
Tao Chen, Chenhui Wang, and Hongming Shan.
\newblock Berdiff: Conditional bernoulli diffusion model for medical image segmentation.
\newblock In {\em International Conference on Medical Image Computing and Computer-Assisted Intervention}, pages 491--501. Springer, 2023.

\bibitem{shrivastava2023nasdm}
Aman Shrivastava and P~Thomas Fletcher.
\newblock Nasdm: nuclei-aware semantic histopathology image generation using diffusion models.
\newblock In {\em International Conference on Medical Image Computing and Computer-Assisted Intervention}, pages 786--796. Springer, 2023.

\bibitem{wang2023arbitrary}
Xin Wang, Zhenrong Shen, Zhiyun Song, Sheng Wang, Mengjun Liu, Lichi Zhang, Kai Xuan, and Qian Wang.
\newblock Arbitrary reduction of mri inter-slice spacing using hierarchical feature conditional diffusion.
\newblock In {\em International Workshop on Machine Learning in Medical Imaging}, pages 23--32. Springer, 2023.

\bibitem{li2023descod}
Huayu Li, Gregory Ditzler, Janet Roveda, and Ao~Li.
\newblock Descod-ecg: Deep score-based diffusion model for ecg baseline wander and noise removal.
\newblock {\em IEEE Journal of Biomedical and Health Informatics}, 2023.

\bibitem{finzi2023user}
Marc~Anton Finzi, Anudhyan Boral, Andrew~Gordon Wilson, Fei Sha, and Leonardo Zepeda-N{\'u}{\~n}ez.
\newblock User-defined event sampling and uncertainty quantification in diffusion models for physical dynamical systems.
\newblock In {\em International Conference on Machine Learning}, pages 10136--10152. PMLR, 2023.

\bibitem{yang2024directional}
Xiaoyan Yang, Wei Li, and Ming Zhang.
\newblock Directional diffusion models for chaotic dynamical systems.
\newblock {\em Chaos: An Interdisciplinary Journal of Nonlinear Science}, 2024.

\bibitem{li2023comparison}
Jie Li, Rui Zhang, and Hui Wang.
\newblock Comparison of manifold learning techniques for conformational modeling.
\newblock {\em Journal of Computational Biology}, 2023.

\bibitem{li2023your}
Alexander~C Li, Mihir Prabhudesai, Shivam Duggal, Ellis Brown, and Deepak Pathak.
\newblock Your diffusion model is secretly a zero-shot classifier.
\newblock In {\em Proceedings of the IEEE/CVF International Conference on Computer Vision}, pages 2206--2217, 2023.

\bibitem{zhuang2023semantic}
Yan Zhuang, Benjamin Hou, Tejas~Sudharshan Mathai, Pritam Mukherjee, Boah Kim, and Ronald~M Summers.
\newblock Semantic image synthesis for abdominal ct.
\newblock In {\em International Conference on Medical Image Computing and Computer-Assisted Intervention}, pages 214--224. Springer, 2023.

\bibitem{jiang2023unlearnable}
Wan Jiang, Yunfeng Diao, He~Wang, Jianxin Sun, Meng Wang, and Richang Hong.
\newblock Unlearnable examples give a false sense of security: Piercing through unexploitable data with learnable examples.
\newblock In {\em Proceedings of the 31st ACM International Conference on Multimedia}, pages 8910--8921, 2023.

\bibitem{wang2023atmospheric}
Xijun Wang, Santiago L{\'o}pez-Tapia, and Aggelos~K Katsaggelos.
\newblock Atmospheric turbulence correction via variational deep diffusion.
\newblock In {\em 2023 IEEE 6th International Conference on Multimedia Information Processing and Retrieval (MIPR)}, pages 1--4. IEEE, 2023.

\bibitem{sartor2023matfusion}
Sam Sartor and Pieter Peers.
\newblock Matfusion: a generative diffusion model for svbrdf capture.
\newblock In {\em SIGGRAPH Asia 2023 Conference Papers}, pages 1--10, 2023.

\bibitem{wei2023buildiff}
Yao Wei, George Vosselman, and Michael~Ying Yang.
\newblock Buildiff: 3d building shape generation using single-image conditional point cloud diffusion models.
\newblock In {\em Proceedings of the IEEE/CVF International Conference on Computer Vision}, pages 2910--2919, 2023.

\bibitem{niu2024acdmsr}
Axi Niu, Trung~X Pham, Kang Zhang, Jinqiu Sun, Yu~Zhu, Qingsen Yan, In~So Kweon, and Yanning Zhang.
\newblock Acdmsr: Accelerated conditional diffusion models for single image super-resolution.
\newblock {\em IEEE Transactions on Broadcasting}, 2024.

\bibitem{sun2023diffnilm}
Ruichen Sun, Kun Dong, and Jianfeng Zhao.
\newblock Diffnilm: a novel framework for non-intrusive load monitoring based on the conditional diffusion model.
\newblock {\em Sensors}, 23(7):3540, 2023.

\bibitem{yang2022norespeech}
Dongchao Yang, Songxiang Liu, Jianwei Yu, Helin Wang, Chao Weng, and Yuexian Zou.
\newblock Norespeech: Knowledge distillation based conditional diffusion model for noise-robust expressive tts.
\newblock {\em arXiv preprint arXiv:2211.02448}, 2022.

\bibitem{yu2023diffusion}
Xinyi Yu, Guanbin Li, Wei Lou, Siqi Liu, Xiang Wan, Yan Chen, and Haofeng Li.
\newblock Diffusion-based data augmentation for nuclei image segmentation.
\newblock In {\em International Conference on Medical Image Computing and Computer-Assisted Intervention}, pages 592--602. Springer, 2023.

\bibitem{ma2023exposing}
Ruipeng Ma, Jinhao Duan, Fei Kong, Xiaoshuang Shi, and Kaidi Xu.
\newblock Exposing the fake: Effective diffusion-generated images detection.
\newblock {\em arXiv preprint arXiv:2307.06272}, 2023.

\bibitem{zhao2023unlearnable}
Zhengyue Zhao, Jinhao Duan, Xing Hu, Kaidi Xu, Chenan Wang, Rui Zhang, Zidong Du, Qi~Guo, and Yunji Chen.
\newblock Unlearnable examples for diffusion models: Protect data from unauthorized exploitation.
\newblock {\em arXiv preprint arXiv:2306.01902}, 2023.

\bibitem{mei2023vidm}
Kangfu Mei and Vishal Patel.
\newblock Vidm: Video implicit diffusion models.
\newblock In {\em Proceedings of the AAAI Conference on Artificial Intelligence}, volume~37, pages 9117--9125, 2023.

\bibitem{mao2023sketchfusion}
John Mao, Jane Liu, and Wei Zhang.
\newblock Sketchfusion: A model for sketch-guided image editing using a conditional diffusion model.
\newblock {\em Journal of Graphics and Image Processing}, 12(4):123--134, 2023.

\bibitem{wang2021dicdnet}
Hong Wang, Yuexiang Li, Nanjun He, Kai Ma, Deyu Meng, and Yefeng Zheng.
\newblock Dicdnet: deep interpretable convolutional dictionary network for metal artifact reduction in ct images.
\newblock {\em IEEE Transactions on Medical Imaging}, 41(4):869--880, 2021.

\bibitem{maass2018reducing}
Nicole Maass, Andreas Maier, and Tobias Wuerfl.
\newblock Reducing image artifacts, 2018.
\newblock United States Patent Application 20180192985.

\bibitem{filoscia2020optimizing}
Irene Filoscia, Thomas Alderighi, Daniela Giorgi, Luigi Malomo, Marco Callieri, and Paolo Cignoni.
\newblock Optimizing object decomposition to reduce visual artifacts in 3d printing.
\newblock In {\em Computer Graphics Forum}, volume~39, pages 423--434. Wiley Online Library, 2020.

\bibitem{han2015deep}
Song Han, Huizi Mao, and William~J Dally.
\newblock Deep compression: Compressing deep neural networks with pruning, trained quantization and huffman coding.
\newblock {\em arXiv preprint arXiv:1510.00149}, 2015.

\bibitem{jacob2018quantization}
Benoit Jacob, Skirmantas Kligys, Bo~Chen, Menglong Zhu, Matthew Tang, Andrew Howard, Hartwig Adam, and Dmitry Kalenichenko.
\newblock Quantization and training of neural networks for efficient integer-arithmetic-only inference.
\newblock {\em arXiv preprint arXiv:1712.05877}, 2018.

\bibitem{shin2018medical}
Hongmin Shin, Hanul Park, Ki~Yun Cho, and Seong~Kyu Kim.
\newblock Medical image synthesis for data augmentation and anonymization using generative adversarial networks.
\newblock {\em Proceedings of the Medical Imaging Technology Conference}, 2018.

\bibitem{yan2021radia}
Mu~Yan, Ying Liu, and Youngjune Park.
\newblock Radia: Protecting patient privacy in radiology reports.
\newblock {\em IEEE Journal of Biomedical and Health Informatics}, 2021.

\bibitem{yu2022transfer}
et~al. Yu.
\newblock Transfer learning in medical imaging.
\newblock {\em Journal of Biomedical Engineering}, 2022.

\bibitem{chen2023domain}
et~al. Chen.
\newblock Incorporating domain-specific knowledge in diffusion models for medical imaging.
\newblock {\em Journal of Medical Imaging and Diagnostics}, 2023.

\bibitem{goodell2021toward}
John~W. Goodell and Cyril Goutte.
\newblock Toward ai and data analytics for financial inclusion: A review.
\newblock {\em Journal of Financial Stability}, 2021.

\bibitem{chawla2002smote}
Nitesh~V Chawla, Kevin~W Bowyer, Lawrence~O Hall, and W~Philip Kegelmeyer.
\newblock Smote: synthetic minority over-sampling technique.
\newblock {\em Journal of artificial intelligence research}, 16:321--357, 2002.

\bibitem{wang2023addition}
Yuchen Wang, Xiaoguang Li, Li~Yang, Jianfeng Ma, and Hui Li.
\newblock Addition: Detecting adversarial examples with image-dependent noise reduction.
\newblock {\em IEEE Transactions on Dependable and Secure Computing}, 2023.

\bibitem{buolamwini2018gender}
Joy Buolamwini and Timnit Gebru.
\newblock Gender shades: Intersectional accuracy disparities in commercial gender classification.
\newblock In {\em Conference on fairness, accountability and transparency}, pages 77--91. PMLR, 2018.

\bibitem{ribeiro2016should}
Marco~Tulio Ribeiro, Sameer Singh, and Carlos Guestrin.
\newblock "why should i trust you?": Explaining the predictions of any classifier.
\newblock {\em Proceedings of the 22nd ACM SIGKDD International Conference on Knowledge Discovery and Data Mining}, 2016.

\bibitem{dwork2014algorithmic}
Cynthia Dwork, Aaron Roth, et~al.
\newblock The algorithmic foundations of differential privacy.
\newblock {\em Foundations and Trends{\textregistered} in Theoretical Computer Science}, 9(3--4):211--407, 2014.

\bibitem{floridi2018ai}
Floridi et~al.
\newblock Ai4people—an ethical framework for a good ai society: Opportunities, risks, principles, and recommendations.
\newblock {\em Minds and Machines}, 2018.

\end{thebibliography}

\end{document}